%% file: main.tex
\documentclass[runningheads]{llncs}

\usepackage{eccv}

\usepackage{eccvabbrv}

\usepackage{graphicx}
\usepackage{booktabs}

\usepackage[accsupp]{axessibility}

\definecolor{eccvblue}{rgb}{0.21,0.49,0.74}
\definecolor{cvprblue}{rgb}{0.21,0.49,0.74}
\usepackage[breaklinks,colorlinks,allcolors=cvprblue]{hyperref}

\usepackage{orcidlink}
\usepackage[table]{xcolor}
\usepackage{capt-of}
\usepackage{float}
\usepackage{makecell}
\usepackage{colortbl}
\usepackage{tcolorbox}
\tcbuselibrary{skins, breakable}
\usepackage{amsmath}
\usepackage{chngcntr}
\usepackage{multirow}
\usepackage{makecell}
\usepackage{colortbl}

\newcommand*\samethanks[1][\value{footnote}]{\footnotemark[#1]}
\begin{document}

\title{InsertAnywhere: Geometrically Grounded and Optics-Aware Video Object Insertion} 

\titlerunning{InsertAnywhere}

\author{Hoiyeong Jin\inst{1}\thanks{Authors contributed equally to this work.}\orcidlink{0009-0002-4653-8888} \and
Hyojin Jang\inst{1}\samethanks\orcidlink{0009-0005-8209-2774} \and
Junha Hyung\inst{1}\samethanks\orcidlink{0009-0002-8244-3659} \and
Jeongho Kim\inst{1}\samethanks\orcidlink{0000-0003-4058-8163} \and
Kinam Kim\inst{1}\orcidlink{0009-0009-5010-0150} \and
Dongjin Kim\inst{1}\orcidlink{0009-0009-5844-0192} \and
Huijin Choi\inst{2}\orcidlink{0009-0005-3208-0513} \and
Hyeonji Kim\inst{2}\orcidlink{0009-0004-8929-1509} \and
Jaegul Choo\inst{1}\orcidlink{0000-0003-1071-4835}}

\authorrunning{H.~Jin, H.~Jang, J.~Hyung, J.~Kim et al.}

\institute{
    KAIST AI \and SK Telecom \\
    \email{\{hy.jin, wkdgywlsrud, rlawjdghek, sharpeeee, kinamplify, dj\_kim, jchoo\}@kaist.ac.kr}\\
    \email{\{astehelen, hyeonji\}@sk.com}\\
}

\maketitle
\newcommand\hy[1]{#1}
\newcommand\hj[1]{#1}
\newcommand\rev[1]{#1}

\begin{center}
  \includegraphics[width=\textwidth]{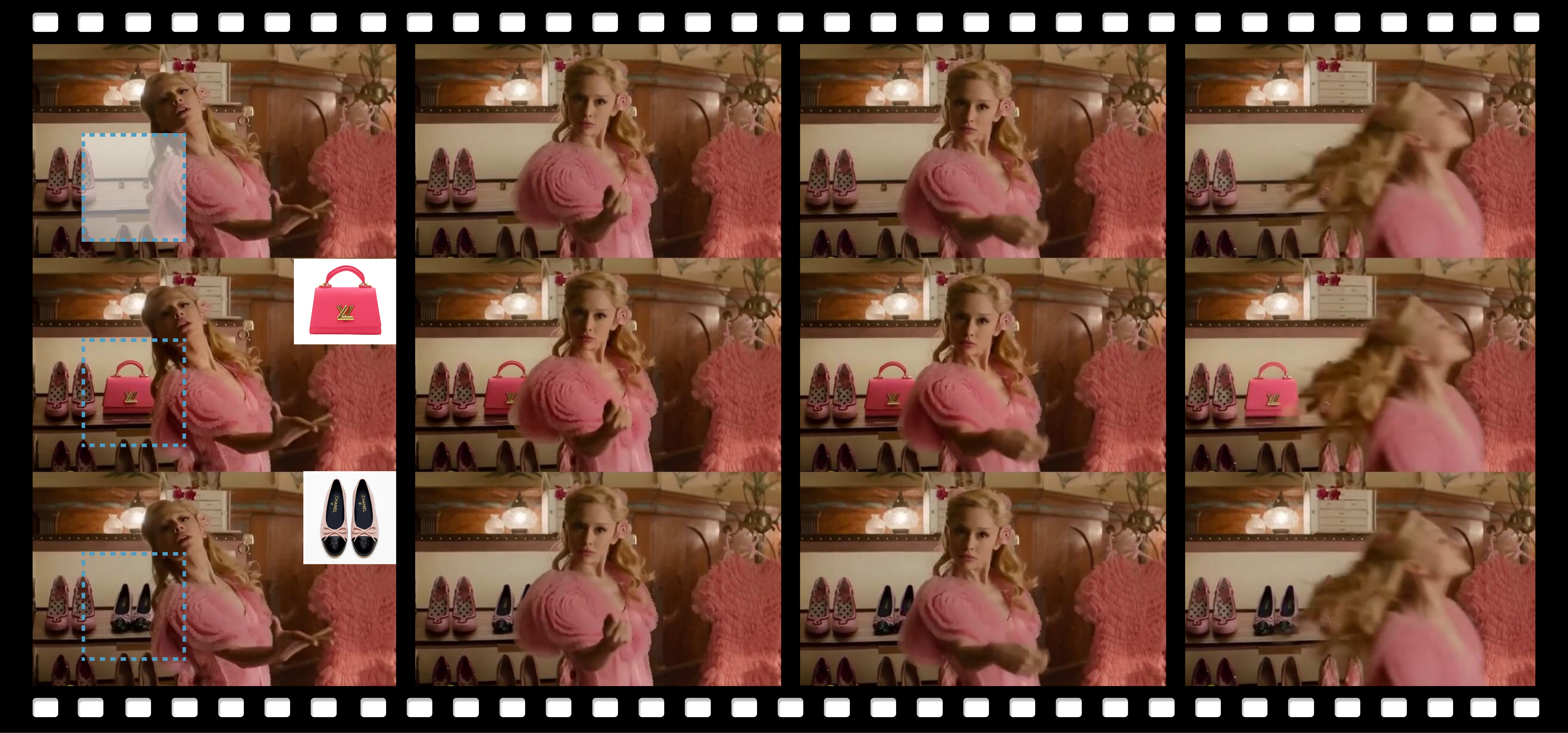}
  \captionof{figure}{\textbf{InsertAnywhere} achieves realistic video object insertion by 4D scene understanding and optics-aware video synthesis. From a single user-specified 3D placement, our method seamlessly propagates the reference object to produce geometrically and photometrically consistent insertions across complex camera motions, occlusions, and viewpoints, while naturally synthesizing optical effects such as shadows and reflections.}
  \label{fig:teaser}
\end{center}

\input{sec/0_abstract}    
\input{sec/1_intro}
\input{sec/2_related_work}

\input{sec/3_method}

\input{sec/4_exp}
\section*{Acknowledgement}
This work was supported by Institute of Information \& Communications Technology Planning \& Evaluation (IITP) grant funded by the Korea government (MSIT) (RS-2019-II190075, Artificial Intelligence Graduate School Program (KAIST)).
This work was supported by Institute of Information \& Communications Technology Planning \& Evaluation (IITP) grant funded by the Korea government (MSIT) (No. RS-2025-02263841, Development of a Real-time Multimodal Framework for Comprehensive Deepfake Detection Incorporating Common Sense Error Analysis).
This work was supported by and carried out in close collaboration with SK Telecom (SKT).
This research was supported by Culture, Sports, and Tourism R\&D Program through the Korea Creative Content Agency (KOCCA) grant funded by the Ministry of Culture, Sports, and Tourism (MCST) in 2024 (Project Name: Development of Technology for Convergence Performance Planning and Production Platform to Revitalize the Production of Convergence Performance by Traditional Artist Dance Music, Project Number: RS-2024-00398536, Contribution Rate: 30\%).

\bibliographystyle{splncs04}
\bibliography{main}
\input{sec/6_suppl}

\end{document}

%% file: sec/0_abstract.tex
\begin{abstract}
Recent advances in diffusion models have enabled impressive video editing capabilities, yet production-grade Video Object Insertion (VOI) remains challenging due to inadequate 4D scene understanding and a lack of proper optical interactions, such as shadows and reflections. To address these limitations, we present InsertAnywhere, a comprehensive VOI framework that achieves geometrically grounded object placement and optics-aware video synthesis. Our approach first leverages a 4D-aware mask generation module that allows users to anchor an object's 3D pose in a single frame. The framework automatically propagates this placement across the video, accurately handling local scene dynamics and occlusions. To synthesize realistic physical lighting interactions, we introduce Optics-Aware Representation Alignment, a novel strategy that utilizes an extended mask to guide feature extraction, enabling optical effects to seamlessly extend beyond the inserted object's boundary. Finally, to overcome the lack of training data for such phenomena, we construct and open-source ROSE++, a specialized quadruplet dataset tailored for the supervised learning of optical effects. Extensive experiments demonstrate that InsertAnywhere produces geometrically plausible and photometrically realistic insertions in complex real-world scenarios, significantly outperforming existing research and commercial generative tools.
\end{abstract}

%% file: sec/1_intro.tex
\section{Introduction}
\label{sec:intro}

Recent advances in diffusion-based generative models have driven significant progress in user-controllable video editing, propelling Video Object Insertion (VOI) into a prominent role for applications like content creation, advertising, and film post-production. The goal of VOI is to seamlessly integrate new objects into existing scenes while maintaining strict spatial, temporal, and photometric consistency. While recent research~\cite{anything,ku2024anyv2v,genprop,tu2025videoanydoor,invi} and commercial video generation tools such as Kling~\cite{Kling} and Pika-Pro~\cite{pikaadditions2025} have made strides in temporally coherent object insertion, existing models still fall short in two critical aspects: 4D-aware object placement and optics-aware video synthesis. These limitations severely restrict their ability to achieve production-grade quality.

Achieving accurate object placement requires both user-guided control and a robust 4D geometric understanding of the scene. Since a single 2D reference image lacks scale and depth context, users must be able to explicitly specify the object's initial position, size, and pose. However, manually annotating this across all video frames is highly impractical. Therefore, a robust framework must automatically propagate the initial placement throughout the sequence while gracefully handling complex real-world dynamics—such as moving support surfaces and occlusions caused by foreground elements—as demonstrated in Fig.~\ref{fig:teaser} and Fig.~\ref{fig:ablation_scene_flow}.

Beyond placement, the generative model must faithfully synthesize the object’s appearance alongside local optical variations induced by the insertion, such as cast shadows and reflections. While mask-free VOI models~\cite{ku2024anyv2v,zi2025senorita,mou2024revideo,genprop,chen2025omniinsert} can hallucinate these effects, they often fail to preserve the unedited background regions. Conversely, mask-conditioned models~\cite{tu2025videoanydoor, zhao2025dreaminsert, jiang2025vace} rigidly confine edits within the provided boundary, struggling to synthesize optical interactions that naturally extend into the surrounding scene. A major bottleneck in solving this is the lack of open-source training data; learning these effects requires a dataset containing a paired tuple of [source video, target video, object mask video, reference image], where the target video explicitly contains the optically integrated object.

To address these challenges, we introduce InsertAnywhere, a comprehensive VOI framework that combines 4D-aware object placement with optics-aware video generation. First, our 4D-aware mask generation module reconstructs the input video into a 4D scene representation. Through an intuitive user interface, users can anchor the object's 3D pose and scale in a single reference frame. Our module then automatically propagates this placement temporally, accurately tracking local scene dynamics (e.g., an object moving synchronously with the luggage cart it rests upon) while preserving occlusion boundaries. The 3D object is then reprojected into 2D to extract a temporally coherent mask video, which serves as the spatial condition for the subsequent video generation stage.

For optics-aware video synthesis, we construct and open-source ROSE++, a synthetic dataset specifically tailored to train models on complex optical interactions. We demonstrate that models fine-tuned on ROSE++ generalize highly effectively to real-world VOI tasks. To maximize visual fidelity, our generation pipeline leverages the strong priors of image-based insertion models via first-frame anchoring. Furthermore, we propose Optics-Aware Representation Alignment to enhance photometric consistency. By aligning the intermediate features of an extended mask with those of the primary input mask, it enables mask-conditioned video diffusion models to synthesize realistic shadows and reflections that seamlessly extend beyond the immediate boundaries of the inserted object.

Our contributions are summarized as follows:
\begin{itemize}
\item We propose a geometrically grounded mask generation module that propagates object masks across all frames via a 4D scene representation. Based on a single user-specified placement, it produces reliable, temporally coherent masks even under complex camera motions and occlusions.
\item We introduce an optics-aware video generation strategy, incorporating a first-frame anchoring technique and Optics-Aware Representation Alignment, which enables the model to synthesize physical lighting interactions, such as shadows and reflections, that extend beyond the object mask.
\item We construct and open-source ROSE++, a specialized dataset for video object insertion comprising quadruplets of source videos, target videos, mask sequences, and reference images, explicitly tailored for the supervised training of optical effects.
\item We demonstrate that \textbf{InsertAnywhere} achieves geometrically plausible and photometrically realistic object insertions across diverse real-world scenarios, significantly outperforming existing commercial and research-based generative tools.
\end{itemize}

%% file: sec/2_related_work.tex
\section{Related Work}

\noindent\textbf{Mask-Free Video Object Insertion}
Mask-free models~\cite{ku2024anyv2v,zi2025senorita,mou2024revideo,genprop,chen2025omniinsert} typically drive insertion using text prompts and an edited first frame. Crucially, the absence of spatial mask conditioning deprives users of the ability to explicitly control the precise location, scale, and boundaries of the inserted object. While their unbounded generation region theoretically allows them to hallucinate natural optical interactions like shadows, this lack of strict spatial constraints frequently degrades unedited background regions. Furthermore, without explicit geometric guidance, these methods struggle to maintain temporal consistency and fail under complex, dynamic occlusion patterns.

\vspace{0.3cm}

\noindent\textbf{Mask-Conditioned Video Object Insertion}
To strictly preserve the background, mask-conditioned models~\cite{invi,chen2024anydoor,tu2025videoanydoor,zhao2025dreaminsert,jiang2025vace} formulate VOI as localized video inpainting governed by an explicit spatial mask. However, this rigid confinement prevents them from synthesizing optical interactions (e.g., cast shadows and reflections) that naturally extend into the surrounding scene. Additionally, these methods typically lack 4D geometric understanding, resulting in brittle and incoherent insertions when the target object interacts with moving support surfaces or becomes occluded.

In contrast to both paradigms, \textbf{InsertAnywhere} achieves production-grade VOI by automatically generating a 4D-aware mask sequence for geometrically grounded, occlusion-robust placement, while simultaneously employing an optics-aware synthesis strategy to render photorealistic lighting interactions that naturally expand beyond the primary object boundary.

%% file: sec/3_method.tex
\begin{figure*}[t]
    \centering
    \includegraphics[width=1.0\textwidth]{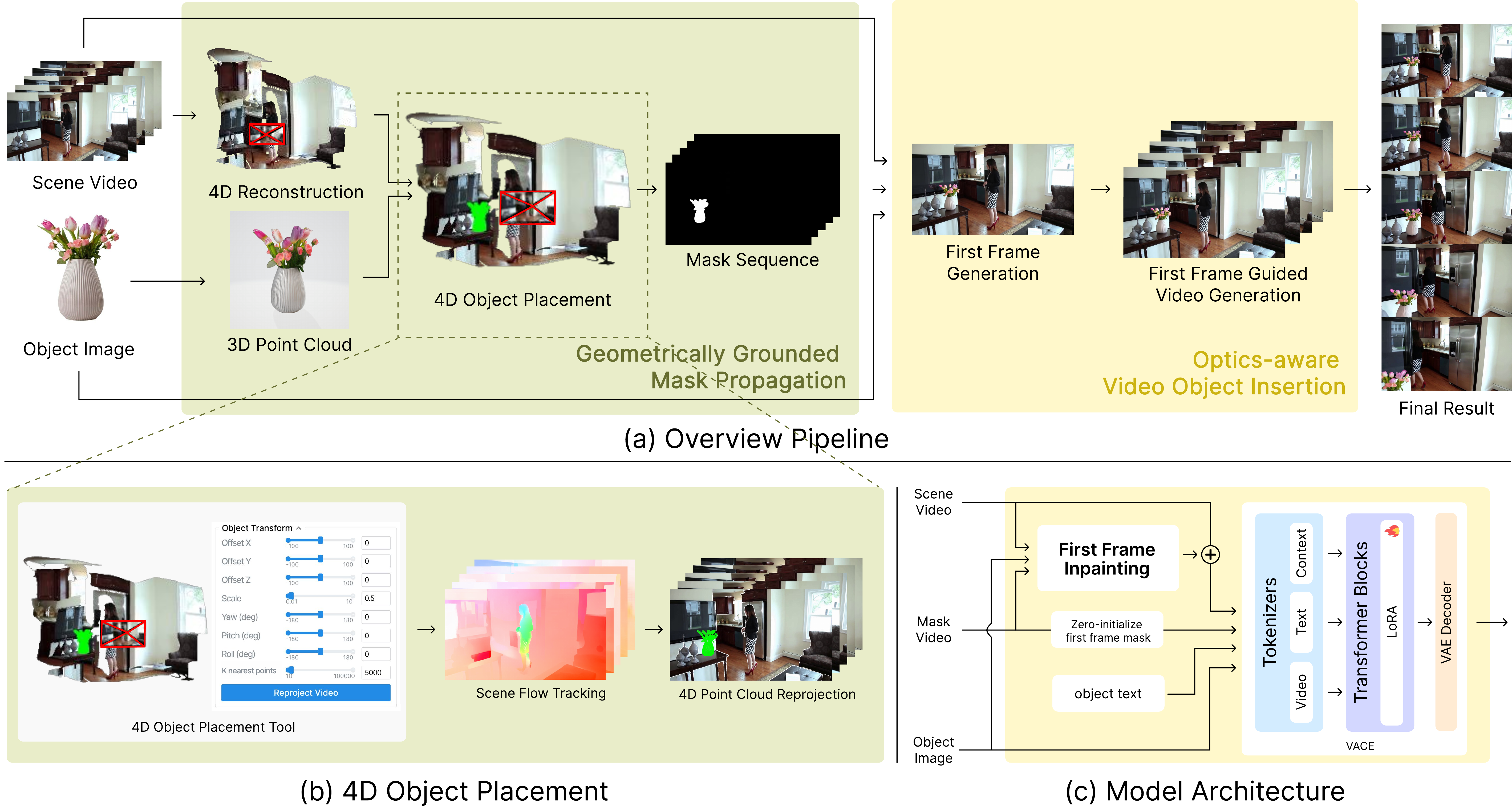}
\caption{The InsertAnywhere architecture. Our framework first utilizes 4D scene reconstruction and scene flow tracking to propagate a single user-specified 3D placement into a geometrically grounded mask sequence (a, b). This mask then conditions an optics-aware diffusion model (c), which leverages first-frame anchoring and LoRA fine-tuning to seamlessly synthesize the inserted object and its surrounding optical effects.}
    \label{fig:overview}
\end{figure*}

\section{Method}

\subsection{Overview}
\label{sec:method:overview}

Video Object Insertion (VOI) is the task of seamlessly integrating a novel object into an existing video sequence. Formally, given a source video $\mathcal{V}_{src} = \{I_1, I_2, \dots, I_T\}$, a reference object image $I_{ref}$, and a user-defined initial 3D placement, the goal is to generate a photorealistic output video $\hat{\mathcal{V}}$ where the object is geometrically and photometrically embedded into the scene.

To achieve production-grade VOI, a framework must overcome two critical limitations present in prior works: the lack of robust 4D scene understanding for controllable object placement, and the rigid, tightly-bound nature of standard mask-conditioned generative models that fail to synthesize scene-wide optical interactions. 

To simultaneously address both spatial controllability and photometric realism, we propose \textbf{InsertAnywhere}. As depicted in \Cref{fig:overview}, our two-stage framework consists of:

\begin{itemize}
    \item \textbf{Geometrically Grounded Mask Propagation (\Cref{sec:method:GGMP}):} Based on a user-specified object placement within a 3D-reconstructed scene, the object is automatically propagated and reprojected using the underlying 4D spatio-temporal dynamics. This yields a reliable, temporally consistent mask sequence $\{M_t\}_{t=1}^{T}$ that accurately handles complex camera motions and occlusions.
    
    \item \textbf{Optics-Aware Video Synthesis (\Cref{sec:method:light-aware_voi}):} Given the propagated mask sequence, source video, and reference image, our diffusion-based generation pipeline synthesizes the final photorealistic video. By leveraging our ROSE++ dataset and a novel Optics-Aware Representation Alignment strategy, the model seamlessly expands object-induced optical effects (e.g., shadows and reflections) beyond the strict boundaries of the primary mask.
\end{itemize}

\subsection{Geometrically Grounded Mask Propagation}
\label{sec:method:GGMP}
This stage generates a highly controllable, geometrically grounded mask sequence based on the source video and a reference object. The pipeline executes three main steps: 4D scene reconstruction, user-controlled object placement, and scene flow-based temporal propagation. This ensures that the resulting mask sequence accurately aligns with the scene's geometry, camera motion, and temporal continuity.

\vspace{0.3cm}

\noindent\textbf{4D Scene Reconstruction.} We first reconstruct a temporally consistent 4D scene representation from the input monocular video by building upon the Uni4D framework~\cite{Yao2025Uni4D}. By orchestrating multiple pretrained vision foundation models including depth estimation~\cite{piccinelli2025unidepthv2}, optical flow~\cite{karaev2025cotracker3}, camera pose estimation, and segmentation~\cite{kirillov2023segment, liu2024grounding, cheng2023tracking}, this stage infers the underlying scene geometry and dynamics. Specifically, the 4D representation consists of spatio-temporal point clouds explicitly decoupled into static components for global structure and dynamic components for moving entities. This decoupling allows our framework to handle complex real-world occlusions and perspective changes with high geometric fidelity.

\vspace{0.3cm}

\noindent\textbf{User-Controlled Object Placement.} To ensure precise spatial control, the target object is introduced into the 4D scene as a standalone rigid entity. First, the reference object image $I_{ref}$ is lifted into a local 3D point cloud $y$ using TRELLIS~\cite{xiang2025structured}, an image-to-3D generation model. Through an interactive interface, users align this object point cloud with the reconstructed scene space by applying a rigid transformation:
\begin{equation}
    y_{s} = s_{obj} R_{obj} y + t_{obj}
    \label{eq:rigid_transform}
\end{equation}
where the rotation $R_{obj}$, translation $t_{obj}$, and global scale factor $s_{obj}$ are intuitively adjusted. This interface provides real-time visualization within the anchor frame $s$, allowing users to verify the object's perspective and scale relative to the supporting scene geometry before committing to video generation.

\vspace{0.3cm}

\noindent\textbf{Scene Flow-Based Object Propagation.}
\label{sec:method:scene_flow}
To maintain physical realism in dynamic scenarios (e.g., an object resting on a moving luggage cart), the object's trajectory must synchronize with local scene dynamics. Simply fixing the initial 3D pose in world coordinates often results in physically implausible drifting relative to moving surfaces. To model these physical interactions, we propagate the object motion using a K-Nearest Neighbor (KNN) scene flow aggregation strategy. 

We first estimate the dense 2D optical flow of scene points around the object using SEA-RAFT~\cite{wang2024sea}. The depth value of each displaced pixel is back-projected into 3D world coordinates using the estimated camera intrinsics and extrinsics, yielding the 3D displacement $\Delta p_{i}^{t}$ for each tracked scene point at frame $t$:
\begin{equation}
    \Delta p_{i}^{t} = p_{i,world}^{t+1} - p_{i,world}^{t}
\end{equation}
For each frame, we select the $K$-nearest scene points to the object centroid in 3D space and aggregate their displacement vectors to estimate the object's motion:
\begin{equation}
    \Delta p_{obj}^{t} = \frac{1}{K} \sum_{k \in \text{TopK}} \Delta p_{k}^{t}
\end{equation}
This aggregated displacement drives the object to faithfully follow the dominant movement of nearby scene surfaces while suppressing noisy flow estimates. To propagate the object in 4D space, the 3D coordinates $y'_{t}$ are sequentially updated:
\begin{equation}
    y'_{t} = y'_{t-1} + \Delta p_{obj}^{t-1}
\end{equation}
where the sequence is initialized with $y'_{s} = y_{s}$ from \Cref{eq:rigid_transform}.

\vspace{0.3cm}

\noindent\textbf{Camera-Aligned Reprojection.} The updated 3D object points $y'_{t}$ are projected onto the image plane of frame $t$ using the estimated camera intrinsics $K$ and extrinsics $P_{t} = [R_{t} | t_{t}]$:
\begin{equation}
    \begin{bmatrix} u_{j,t} \\ v_{j,t} \\ 1 \end{bmatrix} \sim K \left( R_t y'_{j,t} + t_t \right)
\end{equation}
By projecting and rasterizing all visible points, we obtain the object's silhouette for each frame. This reprojection naturally accounts for camera motion, parallax, and occlusion, producing geometrically consistent renderings from real camera viewpoints. The projected silhouettes are further refined using SAM 2~\cite{ravi2024sam} to yield a temporally coherent binary mask sequence $\{M_{t}\}_{t=1}^{T}$, serving as the foundational spatial condition for synthesis.

\begin{figure}[t]
    \centering
    \includegraphics[width=1.0\linewidth]{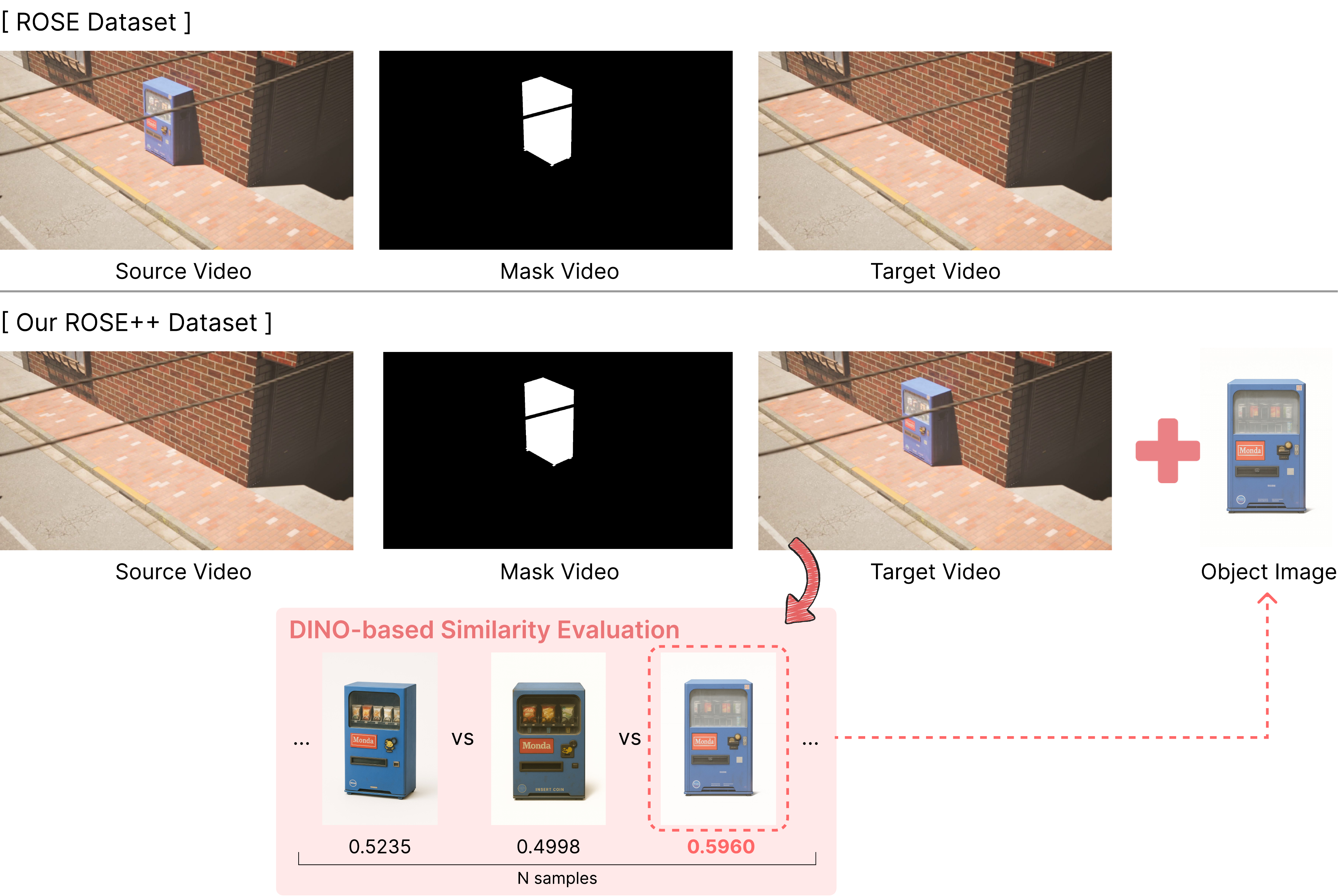}
\caption{ROSE++ dataset generation pipeline. Reference object images are synthesized using an image editing model~\cite{achiam2023gpt} and selected through a DINO-guided rejection sampling pipeline to ensure high appearance fidelity.}
    \label{fig:rose++}
\end{figure}

\subsection{Optics-Aware Video Synthesis}
\label{sec:method:light-aware_voi}
Large-scale mask-conditioned video generation models (e.g., VACE~\cite{jiang2025vace}) provide an efficient foundation for object insertion, as they are trained to strongly adhere to the given spatial condition. However, their localized nature strictly confines generation within the provided mask, leaving the surrounding region untouched. Consequently, they fail to synthesize scene-wide optical interactions, such as cast shadows and reflections, that naturally extend beyond the object's strict boundary. 

\vspace{0.3cm}

\noindent\textbf{ROSE++ Dataset Construction.} To supervise the generation of these physical lighting effects, we construct ROSE++, a specialized dataset derived from the ROSE object removal benchmark~\cite{Miao2025ROSE}. We reorganize the dataset's structure by treating the object-removed video as the source $\mathcal{V}_{src}$ and the object-present video as the target $\mathcal{V}_{tgt}$. This inherently embeds insertion-induced illumination changes into the supervision signal, as shadows and lighting variations absent in the source safely appear in the target.

Since the original dataset lacks isolated reference images, we generate them using an image editing model~\cite{achiam2023gpt} coupled with a rejection sampling pipeline. Multi-view object crops $f_j$ are extracted from $\mathcal{V}_{tgt}$ using the ground-truth masks and provided to the editing model to synthesize candidate reference images on white backgrounds, with scene-dependent lighting cues removed. This ensures that a model trained with ROSE++ infers illumination from the target scene rather than relying on copy-and-paste shortcuts from the reference image.
To filter out candidate images that deviate from the true object's appearance, we employ a rejection sampling strategy. The generated candidates are ranked using a DINO-based similarity metric~\cite{dinov2} against the cropped frames:
\begin{equation}
    s_{k} = \frac{1}{N} \sum_{j=1}^{N} \text{sim}\left(\Phi_{\text{DINO}}(\hat{o}_{k}), \Phi_{\text{DINO}}(f_{j})\right)
\end{equation}
where $\hat{o}_{k}$ denotes the $k$-th generated candidate, $\Phi_{\text{DINO}}(\cdot)$ extracts the feature embeddings, and $\text{sim}(\cdot, \cdot)$ represents cosine similarity. The highest-scoring candidate is selected as the final reference image. 

\vspace{0.3cm}

\begin{figure}[t]
    \centering
    \includegraphics[width=1.0\linewidth]{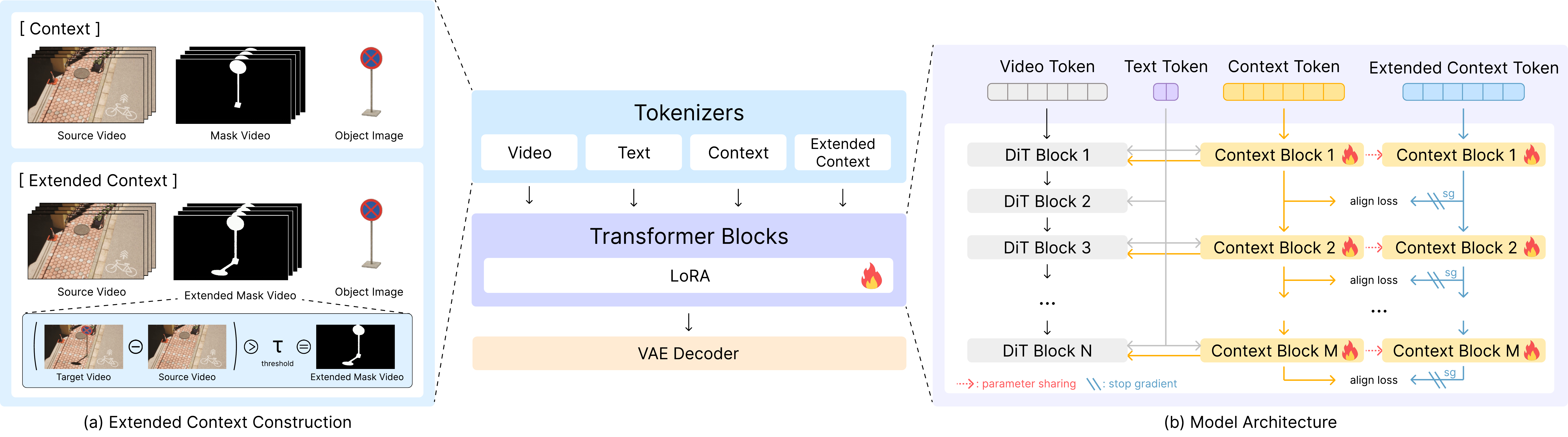}
\caption{(a) We extract an extended mask containing the object and its optical footprint. (b) By explicitly aligning the intermediate representations of the standard fine mask to those of the extended mask via an alignment loss, the model learns to synthesize physical lighting effects (e.g., cast shadows) that extend beyond the strict object mask boundary.}
    \label{fig:arch}
\end{figure}

\noindent\textbf{Optics-Aware Representation Alignment.} 
To encourage the generative model to synthesize optical variations beyond the primary object mask, we propose the representation alignment technique. As illustrated in Fig.~\ref{fig:arch}(a), we compute an \textbf{optics-aware extended mask} ($M_\text{ext}$) by extracting the pixel-wise difference between $\mathcal{V}_{src}$ and $\mathcal{V}_{tgt}$. After thresholding and morphological post-processing, this mask successfully captures both the object region and its associated photometric footprint (i.e., shadows and reflections), such that $M \subset M_\text{ext}$.

During training, both the mask and the extended-mask are tokenized and passed through a shared diffusion backbone (Fig.~\ref{fig:arch}(b)). We extract intermediate features from multiple transformer blocks respectively and enforce alignment between the mask features ($F_{l}$) and the extended-mask features ($F_{l}^{\text{ext}}$):
\begin{equation}
    \mathcal{L}_{align} = \sum_{l} ||F_{l} - \text{sg}(F_{l}^{\text{ext}})||_{2}^{2}
\end{equation}
Crucially, a stop-gradient operator ($\text{sg}$) is applied to the extended-mask branch, allowing it to act as a fixed, optics-aware teacher. This alignment forces the standard fine-mask input to anticipate and generate extended optical variations, even during inference when only the tightly bound fine mask is provided.

\vspace{0.3cm}

\noindent\textbf{First-Frame Anchoring and LoRA Fine-Tuning.} We fine-tune the pretrained mask-conditioned video diffusion model~\cite{jiang2025vace} on ROSE++ using LoRA adapters~\cite{Hu2022LoRA}, preserving its core generative prior while adapting it to optics-aware object insertion. To further maximize visual fidelity, we employ an optional but highly effective first-frame anchoring technique. Because image-level object insertion benefits from highly mature generative priors compared to the more complex VOI task, 
we first synthesize a high-quality edited image using a dedicated image insertion model~\cite{yu2025omnipaint}. We then provide this synthesized image to our video diffusion model as an explicit first-frame conditioning signal. This establishes highly reliable object appearance and local illumination cues, which are naturally propagated to subsequent frames, ensuring robust temporal consistency in color, texture, and lighting throughout the final synthesized video.

%% file: sec/4_exp.tex
\begin{table}[t]
\centering
\resizebox{\linewidth}{!}{
\begin{tabular}{c|c|cc|cccc}
\toprule
& & \multicolumn{2}{c|}{\textbf{Subject Consistency}} & \multicolumn{4}{c}{\textbf{VBench}} \\
\cmidrule(lr){3-4} \cmidrule(lr){5-8}
\textbf{Type} & \textbf{Method}  
& CLIP-I $\uparrow$  
& DINO-I $\uparrow$  
& \makecell{Background \\ Consistency $\uparrow$}
& \makecell{Subject \\ Consistency $\uparrow$}
& \makecell{Motion \\ Smoothness $\uparrow$}
& Imaging Quality $\uparrow$ \\
\midrule  

\multirow{2}{*}{\textbf{Commercial}}
& Pika-Pro~\cite{pikaadditions2025}   
& 0.4940 & 0.3856 & 0.9080 & 0.8720 & 0.9889 & 0.6546 \\  

& Kling~\cite{Kling}      
& 0.6349 & 0.5028 & 0.9335 & 0.9494 & \textbf{0.9940} & 0.7069 \\  

\midrule

\multirow{5}{*}{\textbf{Open Source}}
& AnyV2V~\cite{ku2024anyv2v}
&0.7033 &0.2217 &0.8884 &0.8699 &0.9795 &0.5973  \\

& ReVideo~\cite{mou2024revideo}
&0.7385 &0.3651 &0.9391  &0.9403  &0.9906 &0.6526  \\

& Señorita~\cite{zi2025senorita}
&0.7499  &0.3982  &0.9262  &0.9266  &0.9902 &0.6333 \\

& VACE\textsubscript{our mask}~\cite{jiang2025vace}
& 0.7368 & 0.5060 & 0.9011 & 0.8855 & 0.9887 & 0.6046 \\
\midrule

& Ours\cellcolor[HTML]{DAE8FC}   
& \textbf{0.8132}\cellcolor[HTML]{DAE8FC} 
& \textbf{0.5669}\cellcolor[HTML]{DAE8FC} 
& \textbf{0.9503}\cellcolor[HTML]{DAE8FC}
& \textbf{0.9534}\cellcolor[HTML]{DAE8FC}
& 0.9925\cellcolor[HTML]{DAE8FC}
& \textbf{0.7473}\cellcolor[HTML]{DAE8FC}\\

\bottomrule
\end{tabular}
}
\caption{Quantitative comparisons with baseline methods.}
\label{tab:quantitative_comparison}
\end{table}

\section{Experiments}
\subsection{Experimental Setup}
\noindent\textbf{Evaluation Dataset. }
We introduce VOIBench, a new benchmark designed to rigorously assess video object insertion.
\rev{VOIBench spans three axes that jointly stress geometric and photometric robustness: \emph{scene} (indoor / outdoor / natural), \emph{occlusion} (none / dynamic / initially-occluded), and \emph{lighting} (daylight / indoor / low-light), giving the benchmark broad coverage and evaluative power.}
The dataset consists of \rev{200} video clips covering a broad spectrum of settings, such as indoor scenes, outdoor environments, and natural landscapes, with each video containing a pair of objects.
We crawled contextually relevant objects for each scene. 

\vspace{0.3cm}

\noindent\textbf{Baselines and Evaluation Metrics. }
We evaluate against top commercial tools (Pika-Pro~\cite{pikaadditions2025}, Kling~\cite{Kling}) and representative open-source models (AnyV2V \cite{ku2024anyv2v}, ReVideo~\cite{mou2024revideo}, Señorita~\cite{zi2025senorita}), as well as a VACE-based variant (VACE$_{\text{our mask}}$), conditioned on our generated mask, to ensure fair VOI evaluation.
We also report ablation results to isolate the impact of our proposed components.

We assess three key aspects: (1) \textit{Subject Consistency} via CLIP-I~\cite{clip} and DINO-I~\cite{dinov2}; (2) \textit{Video Quality} using VBench~\cite{vbench++} (Image Quality, Background/ Subject Consistency, Motion Smoothness); and (3) \textit{Multi-View Consistency}~\cite{vbench++} to gauge how reliably the object is preserved across viewpoint changes and dynamic occlusions.

\begin{figure*}[t]
    \centering
    \includegraphics[width=1.0\textwidth]{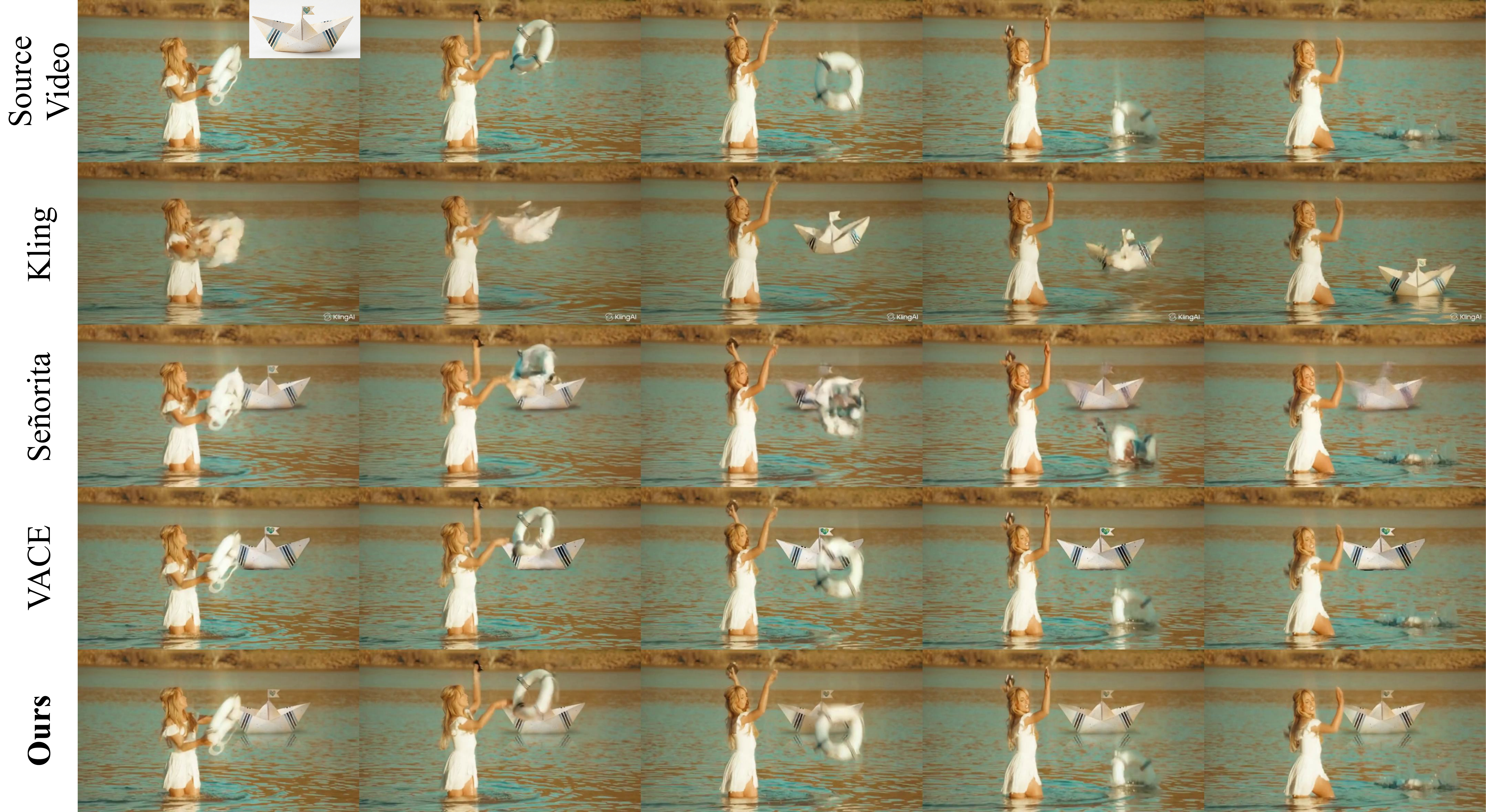}
    \caption{Qualitative comparisons with baseline methods. Best viewed when zoomed in.}
    \label{fig:qual_result}
\end{figure*}

\subsection{Qualitative Results}
Fig.~\ref{fig:qual_result} qualitatively highlights two key factors: 1) a 4D-aware mask for occlusion-robust insertion and 2) Optics-Aware Representation Alignment for realistic illumination effects.
Without a 4D-aware mask, Kling~\cite{Kling} and Señorita~\cite{zi2025senorita} fail to maintain a stable separation between the insertion region and the rest of the scene, leading to background drift and deformation when a dynamic occluder (the moving tube) crosses the target area.
In particular, even with a detailed prompt specifying the insertion location (``Add a paper boat to the right rear of the river in the video background.''), Kling often misinterprets the scene and transforms the tube into a paper boat.
Señorita, which propagates edits from an edited first frame, degrades over time and cannot sustain consistent insertions under occlusions.
In contrast, methods equipped with a 4D-aware mask (VACE~\cite{jiang2025vace} and ours) preserve both the tube geometry and the original background, producing consistent insertions.
However, VACE lacks optics-aware modeling, making the inserted boat appear pasted with implausible reflections and weak cast shadows.
Additional qualitative results for other baselines are provided in the supplementary material.

\subsection{Quantitative Results}
To assess subject consistency, we uniformly sample 10 frames from each generated video and measure how closely they match the reference subject using subject-masked regions only.
As shown in Table~\ref{tab:quantitative_comparison}, InsertAnywhere achieves the highest CLIP-I~\cite{clip} and DINO-I~\cite{dinov2} scores among all compared methods, demonstrating superior identity preservation.
Moreover, InsertAnywhere attains the best overall VBench score, with clear advantages in Background Consistency, Subject Consistency, and Imaging Quality. Overall, the results suggest that our model not only provides a reliable 4D-aware mask for subject control but also improves full-video quality without sacrificing visual fidelity.
In contrast, Kling and Pika-Pro often introduce undesirable changes to the original background during insertion.
\rev{To verify that our gains stem from our design rather than from data exposure, we additionally fine-tune VACE on ROSE++ with our 4D-aware mask and with a coarse bounding-box mask (Table~\ref{tab:fair_vace}). Both variants underperform our full model, confirming that the improvement comes from our optics-aware design rather than merely from training on ROSE++.}
\rev{Notably, our method is best on five of the six metrics; the only exception is Motion Smoothness, where we closely match Kling ($0.9925$ vs.\ $0.9940$), a gap within VBench's per-clip noise.}

\begin{table}[t]
\centering
\resizebox{\linewidth}{!}{%
\begin{tabular}{l|cc|cccc}
\toprule
& \multicolumn{2}{c|}{\textbf{Subject Consistency}} & \multicolumn{4}{c}{\textbf{VBench}} \\
\cmidrule(lr){2-3} \cmidrule(lr){4-7}
\textbf{Method}
 & CLIP-I $\uparrow$
 & DINO-I $\uparrow$
 & \makecell{Background \\ Consistency $\uparrow$}
 & \makecell{Subject \\ Consistency $\uparrow$}
 & \makecell{Motion \\ Smoothness $\uparrow$}
 & Imaging Quality $\uparrow$ \\
\midrule
VACE\textsubscript{ - , our 4D mask}~\cite{jiang2025vace}     & 0.7368 & 0.5060 & 0.9011 & 0.8855 & 0.9887 & 0.6046 \\
VACE\textsubscript{ROSE++, bbox mask}~\cite{jiang2025vace}    & 0.6383 & 0.4856 & 0.9026 & 0.8818 & 0.9864 & 0.5824 \\
VACE\textsubscript{ROSE++, our 4D mask}~\cite{jiang2025vace}  & 0.7485 & 0.5103 & 0.9035 & 0.8895 & 0.9888 & 0.6583 \\
\textbf{Ours\cellcolor[HTML]{DAE8FC}}    & \textbf{0.8132\cellcolor[HTML]{DAE8FC}} & \textbf{0.5669\cellcolor[HTML]{DAE8FC}} & \textbf{0.9503\cellcolor[HTML]{DAE8FC}} & \textbf{0.9534\cellcolor[HTML]{DAE8FC}} & \textbf{0.9925\cellcolor[HTML]{DAE8FC}} & \textbf{0.7473\cellcolor[HTML]{DAE8FC}} \\
\bottomrule
\end{tabular}}
\caption{\rev{Fair comparison against VACE fine-tuned on ROSE++ with our 4D-aware mask and with a coarse bounding-box mask, evaluated on the VOIBench. Both fine-tuned VACE variants underperform our full model.}}
\label{tab:fair_vace}
\end{table}

\begin{figure}[t]
    \centering
    \includegraphics[width=1.0\linewidth]{figures/2026eccv_abl.pdf}
    \caption{Qualitative ablation study. Sequentially adding our proposed components progressively improves occlusion handling, object fidelity, and realistic optics-aware generation. Best viewed zoomed in.}
    \label{fig:abl_qual}
\end{figure}

\subsection{Ablation Study}
\begin{figure}[t]
    \centering
    \includegraphics[width=1.0\linewidth]{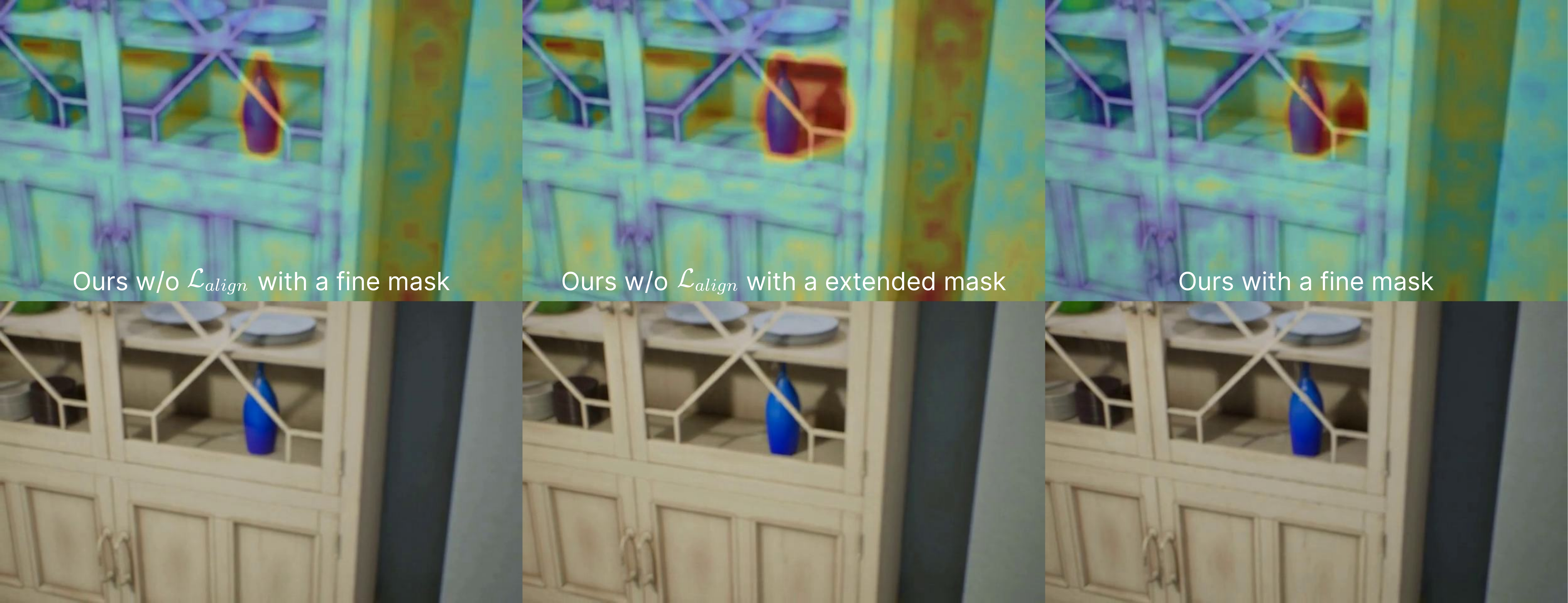}
    \caption{Impact of Optics-Aware Representation Alignment. Attention maps for (a) a baseline model (w/o $\mathcal{L}_{align}$) with a fine mask, (b) a baseline model with an extended mask, and (c) our model with a fine mask. Despite training on the same ROSE++ dataset, the baseline (a) remains strictly confined to the inserted object. In contrast, our aligned model (c) exhibits strong external activations closely matching (b). This confirms that our alignment strategy effectively trains the network to synthesize surrounding optical effects (e.g., shadows) without requiring an extended input mask.}
    \label{fig:abl_oara_attn}
\end{figure}

We evaluate the effectiveness of our proposed components through a progressive ablation study, with qualitative and quantitative results detailed in Fig.~\ref{fig:abl_qual} and Tab.~\ref{tab:quantitative_comparison_abl}. First, introducing our 4D-aware mask sequence successfully resolves dynamic occlusion failures (e.g., orange box, row 2), though basic object fidelity remains low. Next, incorporating first-frame anchoring significantly enhances identity preservation, yet temporal inconsistencies in shape and color still emerge post-occlusion. While subsequent LoRA fine-tuning on the ROSE++ dataset successfully stabilizes temporal consistency and appearance, the network still struggles to hallucinate external optical interactions. Finally, integrating our Optics-Aware Representation Alignment ($\mathcal{L}_{align}$) naturally synthesizes insertion-induced lighting effects, such as shadows. Combining all components yields geometrically accurate and photometrically realistic insertions with high object fidelity across the entire sequence.

\begin{figure}[t]
    \centering
    \includegraphics[width=1.0\linewidth]{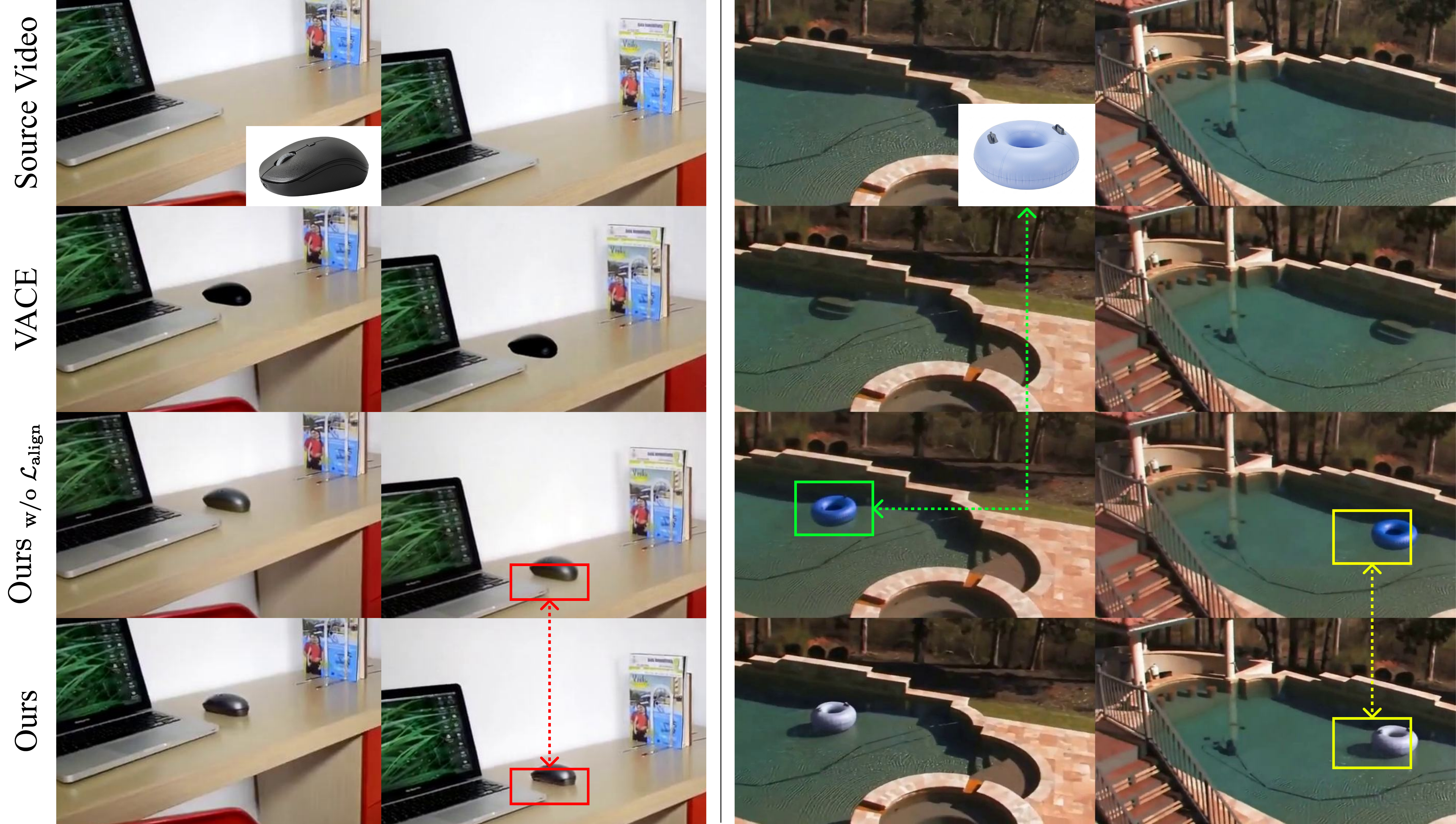}
\caption{Qualitative ablation for Optics-Aware Representation Alignment.}
    \label{fig:abl_oara}
\end{figure}

\begin{table*}[t!]
\centering
\resizebox{\textwidth}{!}{
\begin{tabular}{l|cc|cccc|c}
\toprule
& \multicolumn{2}{c|}{\textbf{Subject Consistency}} & \multicolumn{5}{c}{\textbf{VBench}} \\
\cmidrule(lr){2-3} \cmidrule(lr){4-7} \cmidrule(lr){7-8}
\textbf{Method}  
& CLIP-I $\uparrow$  
& DINO-I $\uparrow$  
& \makecell{Background \\ Consistency $\uparrow$}
& \makecell{Subject \\ Consistency $\uparrow$}
& \makecell{Motion \\ Smoothness $\uparrow$}
& Imaging Quality $\uparrow$ 
& \makecell{Multi-View \\ Consistency $\uparrow$} \\ 
\midrule

Config (a)   
&0.7532   &0.3861   &0.9232   &0.9380   &0.9913 &0.6298  &0.5308   \\ 

Config (b)   
& 0.7880 & 0.5135 &0.9175   &0.9290   &0.9911  &0.6318  &0.5436   \\ 

Config (c)
& 0.8122 & \textbf{0.5678} &0.9429   &0.9520   &0.9916  &0.7101  &0.5857   \\

Ours \cellcolor[HTML]{DAE8FC}
& \textbf{0.8132}\cellcolor[HTML]{DAE8FC}
& 0.5669\cellcolor[HTML]{DAE8FC}
& \textbf{0.9503}\cellcolor[HTML]{DAE8FC}
& \textbf{0.9534}\cellcolor[HTML]{DAE8FC}
& \textbf{0.9925}\cellcolor[HTML]{DAE8FC}
& \textbf{0.7473}\cellcolor[HTML]{DAE8FC}
& \textbf{0.5865}\cellcolor[HTML]{DAE8FC}  \\

\bottomrule
\end{tabular}
}
\caption{Quantitative results of the ablation study.}
\label{tab:quantitative_comparison_abl}
\end{table*}

\noindent\textbf{Effect of Alignment Loss on Optics-Aware Rendering.}
Table~\ref{tab:mv_consistency_and_shadow} (right) reports quantitative comparisons for rendering quality using LPIPS/PSNR/SSIM against the ground truth. To enable a paired evaluation with ground truth, we hold out 25 samples from ROSE++ as a test set and compute all metrics on this split. We compare VACE, our ROSE++ LoRA fine-tuned model without the alignment loss (Ours \textsubscript{w/o$\mathcal{L}_{\mathrm{align}}$} ), and the full model trained with the proposed alignment loss (Ours). We evaluate results under two complementary regions: Object region covers only the precise foreground area of the inserted object, while Optics region isolates the area affected by the object-induced optical effects such as shadows, specular highlights, and contact-dependent illumination changes. Using Object Mask, the two Ours variants achieve similar performance, indicating that the object appearance fidelity is largely insensitive to $\mathcal{L}_{align}$. In contrast, under Optics Mask, Ours yields clear gains with higher PSNR/SSIM and lower LPIPS, demonstrating that $\mathcal{L}_{align}$ effectively improves the preservation and consistency of optics-aware effects.
\rev{We further verify that $\mathcal{L}_{align}$ does not degrade fidelity in the unedited region: across the held-out ROSE++ split, the unedited-region PSNR is nearly identical with and without $\mathcal{L}_{align}$ (VACE\textsubscript{our mask}: 33.27, Ours\textsubscript{w/o $\mathcal{L}_{align}$}: 33.31, Ours: 33.30), while the optics-region PSNR rises by $+7.3$\,dB. The mild color drift occasionally observed thus reflects sample-to-sample variation rather than a systematic effect of $\mathcal{L}_{align}$.}
Qualitative results in Fig. ~\ref{fig:abl_oara} corroborate this trend: in both the desk-top mouse and the floating tube examples, VACE shows noticeable appearance mismatch to the reference and reduced overall image quality, whereas both Ours variants maintain comparable object quality, with Ours producing more realistic and coherent specular highlights and shadows than Ours \textsubscript{w/o $\mathcal{L}_{\mathrm{align}}$}.

Fig.~\ref{fig:abl_oara_attn} visualizes the network's attention maps during generation. We compare three configurations: (a) a baseline model trained without $\mathcal{L}_{align}$ conditioned on a fine mask, (b) baseline model conditioned on an extended mask, and (c) our full model conditioned on a fine mask. Despite being fine-tuned on the same ROSE++ dataset, the attention of the baseline model (a) remains strictly confined to the inserted object's boundaries. In contrast, when provided only the tightly-bound fine mask, our aligned model (c) exhibits strong external activations that closely mirror the extended mask conditions of (b). This visual evidence confirms that our alignment strategy effectively teaches the network to anticipate and synthesize surrounding optical effects, such as cast shadows, without requiring an explicitly expanded mask during inference.

\begin{table}[t!]
\centering

\begin{minipage}[t]{0.32\columnwidth}
\centering
\resizebox{0.95\columnwidth}{!}{%
\begin{tabular}{cc}
\toprule
Methods & \makecell{Multi-View \\ Consistency $\uparrow$} \\ \hline
Ours\textsubscript{random} & 0.5295 \\
Ours\cellcolor[HTML]{DAE8FC} & \textbf{0.5865}\cellcolor[HTML]{DAE8FC} \\
\bottomrule
\end{tabular}%
}
\end{minipage}
\hfill
\begin{minipage}[t]{0.67\columnwidth}
\centering
\resizebox{0.95\columnwidth}{!}{%
\begin{tabular}{c|ccc|ccc}
\toprule
& \multicolumn{3}{c|}{\makecell{Object Mask}} 
& \multicolumn{3}{c}{\makecell{Optics Mask}} \\
Methods 
& \makecell{LPIPS $\downarrow$} & PSNR $\uparrow$ & SSIM $\uparrow$  
& \makecell{LPIPS $\downarrow$} & PSNR $\uparrow$ & SSIM $\uparrow$\\ 
\hline
VACE\textsubscript{our mask}~\cite{jiang2025vace} & 0.0250 & 21.1980  & 0.9795 & 0.0130 & 19.8345 & 0.9877\\
Ours \textsubscript{w/o $\mathcal{L}_{\mathrm{align}}$} 
& \textbf{0.0137}\cellcolor[HTML]{DAE8FC} & \textbf{26.0282}\cellcolor[HTML]{DAE8FC}  & 0.9843 
& 0.0121 & 19.9065 & 0.9883 \\
Ours\cellcolor[HTML]{DAE8FC} 
& 0.0161 & 25.8785 & \textbf{0.9872}\cellcolor[HTML]{DAE8FC} 
& \textbf{0.0089}\cellcolor[HTML]{DAE8FC} & \textbf{27.2362}\cellcolor[HTML]{DAE8FC} & \textbf{0.9920}\cellcolor[HTML]{DAE8FC} \\
\bottomrule
\end{tabular}%
}
\end{minipage}

\caption{Left: Multi-view Consistency evaluation. Models trained with random-frame references vs. our ROSE++ reference. Right: Quantitative evaluation of shadow rendering quality on object and optics masks, comparing VACE, Ours\textsubscript{w/o $\mathcal{L}_{\mathrm{align}}$}, and Ours.}
\label{tab:mv_consistency_and_shadow}
\vspace{-0.5cm}
\end{table}

\vspace{0.3cm}

\noindent\textbf{Effectiveness of ROSE++}
We evaluate whether the generated reference images in ROSE++ successfully prevent copy-and-paste artifacts. To do so, we compare our model against a baseline trained with reference objects trivially cropped from the target video. As reported in Table~\ref{tab:mv_consistency_and_shadow} (left), our method achieves substantially higher multi-view consistency. This confirms that our dataset construction forces the network to genuinely infer 4D geometry and lighting, rather than relying on direct image copying.

\begin{figure}[t]
    \centering
    \includegraphics[width=1.0\linewidth]{figures/2026eccv_harry.pdf}
\caption{Effectiveness of scene flow-based object propagation in~\Cref{sec:method:scene_flow}. Unlike the static mask (second row), our proposed propagation ensures temporally consistent and 4D-aware alignment with moving objects (third row).}
    \label{fig:ablation_scene_flow}

\end{figure}

\section{Conclusion}
We introduced \textbf{InsertAnywhere}, a novel framework for production-grade Video Object Insertion (VOI). By combining Geometrically Grounded Mask Propagation with an optics-aware diffusion model, our approach simultaneously guarantees precise spatial control and photorealistic rendering. From a single 3D anchor, our method automatically handles complex camera motions and dynamic occlusions. Furthermore, by leveraging our custom ROSE++ dataset and Optics-Aware Representation Alignment, the network successfully synthesizes object-induced physical lighting effects, such as shadows, outside the immediate mask boundary. Extensive evaluations confirm that InsertAnywhere significantly outperforms existing baselines in both geometric consistency and photometric realism, paving the way for advanced practical applications in virtual product placement and visual effects.

\rev{\noindent\textbf{Limitations.} Even when the propagated mask is not perfectly pixel-accurate, our pipeline re-synthesizes regions beyond the boundary and absorbs small errors into a plausible output. Nonetheless, complex physics-aware interactions (\eg, dropping a heavy object onto a soft sofa) lie outside our current scope and are left for future work.}

%% file: sec/6_suppl.tex
\clearpage
\appendix
\setcounter{page}{1}
\renewcommand{\thesection}{\Alph{section}}

\renewcommand{\thefigure}{\thesection\arabic{figure}}
\renewcommand{\thetable}{\thesection\arabic{table}}

\setcounter{figure}{0}
\setcounter{table}{0}

{\centering\Large\vspace{0.5em}
\textbf{InsertAnywhere: Geometrically Grounded and Optics-Aware Video Object Insertion} \\
}

{\centering\large
Supplementary Material \\
\vspace{1.0em}
}

This supplementary material covers:
Implementation Details (Sec.~\ref{sec:imple}),
User Study (Sec.~\ref{sec:userstudy}),
Retrieval Prompt Details (Sec.~\ref{sec:retreival}),
and Additional Qualitative Comparisons (Sec.~\ref{sec:add_qual}).

\section{Implementation Detail.}
\label{sec:imple}
\noindent\textbf{Training Details.}
Our video generation is performed with Wan2.1-VACE-14B~\cite{jiang2025vace}, a diffusion-based video generation model fine-tuned with LoRA to adapt to our insertion-specific domain. 
The model is fine-tuned for 6{,}500 iterations with a learning rate of $1\times10^{-4}$ and a LoRA rank of 128. 
The entire training process, conducted on a single NVIDIA H200 GPU, takes approximately 60 hours.
All videos are trained and generated at a spatial resolution of $832\times480$ with 81 frames per clip.

\noindent\textbf{Inference Cost.} On a single NVIDIA B200 GPU, our pipeline takes about 4 minutes for the 4D-mask stage and 4.5 minutes for video synthesis at $832\times480\times81$ (peak memory 48\,GB), comparable to VACE-14B (4.5 minutes) and orders of magnitude faster than manual graphics pipelines. As VOI is an offline production task, this footprint is well within practical limits.

\rev{\noindent\textbf{Extended-mask noise.}
Because ROSE is simulator-rendered, the noise in the extended (optics) mask is small. We further reduce it by frame-differencing the source and target edits, retaining only connected components near the ground-truth object mask by Euclidean-distance-transform (EDT) proximity, and applying morphological closing to merge the object with its associated shadow.}

\section{User Study} 
\label{sec:userstudy}

\begin{figure*}[t]
    \centering
    \includegraphics[width=1.0\textwidth]{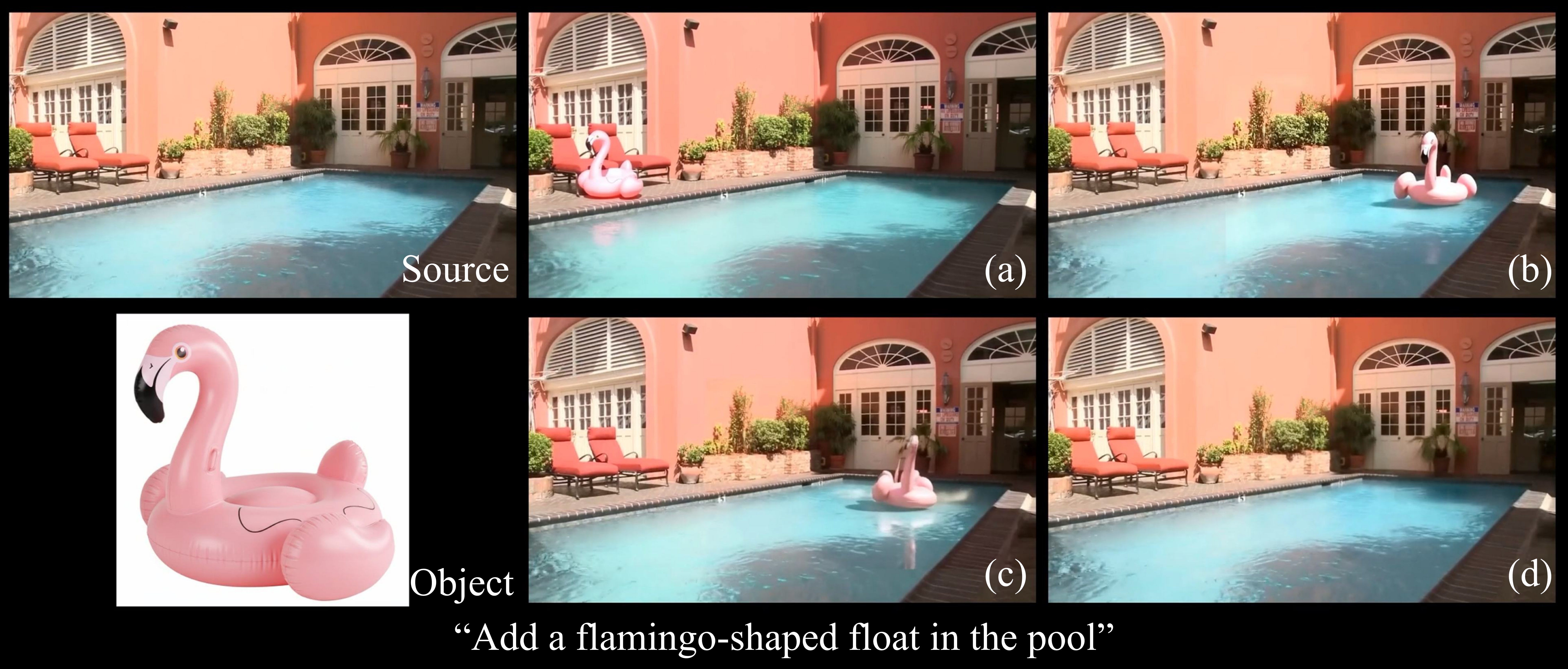}
\caption{User study example}
    \label{fig:user_study}
\end{figure*}

We conducted a user study with 20 participants by randomly sampling 10\% of the videos in the test set. For each video, we prepared six evaluation questions, and participants were asked to choose the best-performing model for each question from four candidates: our method and three baselines selected based on the top scores in Table~\ref{tab:quantitative_comparison}, namely Kling from the commercial group and Señorita and VACE\textsubscript{our mask} from the open-source group. To avoid positional bias, the order of the four candidate results was randomly shuffled for each participant. As illustrated in the Fig.~\ref{fig:user_study}, each question presented the source video and the reference object together with four candidate results, labeled (a)-(d), whose ordering was randomized across participants.

The evaluation questions are as follows:

\begin{itemize}
    \item \textbf{Object Realism.} Assesses whether the inserted object exhibits physically plausible geometry, scale, and appearance without distortions or implausible configurations.

    \item \textbf{Lighting Consistency.} Measures how well the inserted object matches the illumination and shading conditions of the surrounding scene.

    \item \textbf{Occlusion Integrity.} Evaluates whether occlusion relationships with surrounding objects are handled correctly, without causing disappearance or distortion of existing scene elements.

    \item \textbf{Object--Video Consistency.} Determines how consistent the inserted object is with visually or semantically related objects present in the original video.

    \item \textbf{Background Preservation.} Checks whether regions unrelated to the inserted object remain faithful to the original video without unnecessary alterations or artifacts.

    \item \textbf{Overall Naturalness.} Captures the overall perceptual realism of the result, including the absence of visible artifacts and the user’s preferred choice for real-world deployment.
\end{itemize}

Table~\ref{tab:user_study_table} reports the user preference percentages aggregated over the randomly sampled 10\% subset of the test videos and all 20 participants. For each evaluation criterion, participants selected the method they judged to perform best, and the percentages indicate the proportion of total votes received by each method for that criterion.

\begin{table}[t]
\centering
\resizebox{1.0\linewidth}{!}{
\begin{tabular}{l|cccccc}
\toprule
Method (\%)
& \makecell{Object \\ Realism} 
& \makecell{Lighting \\ Consistency} 
& \makecell{Occlusion \\ Integrity} 
& \makecell{Object-Video \\ Consistency} 
& \makecell{Background \\ Preservation} 
& \makecell{Overall \\ Naturalness} \\
\midrule

Kling~\cite{Kling}    
& 19.09 & 34.55 & 13.64 & 20.45 & 11.82 & 21.82 \\

Señorita~\cite{zi2025senorita}    
& 6.82 & 15.45 & 6.36 & 7.73 & 10.91 & 10.91 \\

VACE\textsubscript{our mask}~\cite{jiang2025vace}   
& 10.45 & 13.64 & 38.18 & 11.36 & 37.27 & 8.64 \\

Ours\cellcolor[HTML]{DAE8FC}  
& \textbf{63.64}\cellcolor[HTML]{DAE8FC} 
& \textbf{36.36}\cellcolor[HTML]{DAE8FC}
& \textbf{41.82}\cellcolor[HTML]{DAE8FC}
& \textbf{60.45}\cellcolor[HTML]{DAE8FC}
& \textbf{40.00}\cellcolor[HTML]{DAE8FC}
& \textbf{58.64}\cellcolor[HTML]{DAE8FC} \\
\bottomrule
\end{tabular}
}
\caption{User study preference percentages across six evaluation criteria for four compared methods.}

\label{tab:user_study_table}
\end{table}

\rev{\noindent\textbf{Optics-Focused User Study.} To more directly assess the perceptual benefit of our optics-aware synthesis, we conducted an additional user study with 30 participants on 50 clips explicitly selected for shadow- and reflection-rich scenes. As reported in Table~\ref{tab:user_study_optics}, our method is again preferred across all six criteria, confirming that the optics-aware gain, not only the overall preference, is perceptually salient.}

\begin{table}[t]
\centering
\resizebox{1.0\linewidth}{!}{
\begin{tabular}{l|cccccc}
\toprule
Method (\%)
& \makecell{Object \\ Realism}
& \makecell{Lighting \\ Consistency}
& \makecell{Occlusion \\ Integrity}
& \makecell{Object-Video \\ Consistency}
& \makecell{Background \\ Preservation}
& \makecell{Overall \\ Naturalness} \\
\midrule

Kling~\cite{Kling}
& 24.5 & 26.0 & 12.0 & 16.4 & 9.7 & 15.0 \\

Señorita~\cite{zi2025senorita}
& 7.9 & 22.7 & 6.0 & 6.7 & 8.9 & 8.5 \\

VACE\textsubscript{our mask}~\cite{jiang2025vace}
& 9.7 & 5.3 & 26.0 & 10.3 & 28.5 & 18.5 \\

VACE\textsubscript{ROSE++, our mask}~\cite{jiang2025vace}
& 14.2 & 14.0 & 27.0 & 25.5 & 21.5 & 23.0 \\

Ours\cellcolor[HTML]{DAE8FC}
& \textbf{43.7}\cellcolor[HTML]{DAE8FC}
& \textbf{32.0}\cellcolor[HTML]{DAE8FC}
& \textbf{29.0}\cellcolor[HTML]{DAE8FC}
& \textbf{41.1}\cellcolor[HTML]{DAE8FC}
& \textbf{31.5}\cellcolor[HTML]{DAE8FC}
& \textbf{35.0}\cellcolor[HTML]{DAE8FC} \\
\bottomrule
\end{tabular}
}
\caption{\rev{Optics-focused user study with 30 participants on 50 shadow/reflection-rich clips. Our method is preferred across all six criteria.}}
\label{tab:user_study_optics}
\end{table}

\section{ROSE++ Prompt Details.}
\label{sec:retreival}
We use the prompt described in Fig.~\ref{fig:gpt_prompt} to generate object images for constructing the ROSE++ dataset. 
For each video, we use this prompt to generate multiple object candidates with consistent visual appearance and select as the reference image the one with the highest DINO score.

\begin{figure}[t]
    \centering
    \includegraphics[width=1.0\textwidth]{figures/2026eccv_rose_prompt.pdf}
    \caption{Prompt for object retrieval in constructing the ROSE++ dataset}
    \label{fig:gpt_prompt}
\end{figure}

\section{Additional Qualitative Comparisons.}
\label{sec:add_qual}

\begin{figure}[t]
    \centering
    \includegraphics[width=1.0\textwidth]{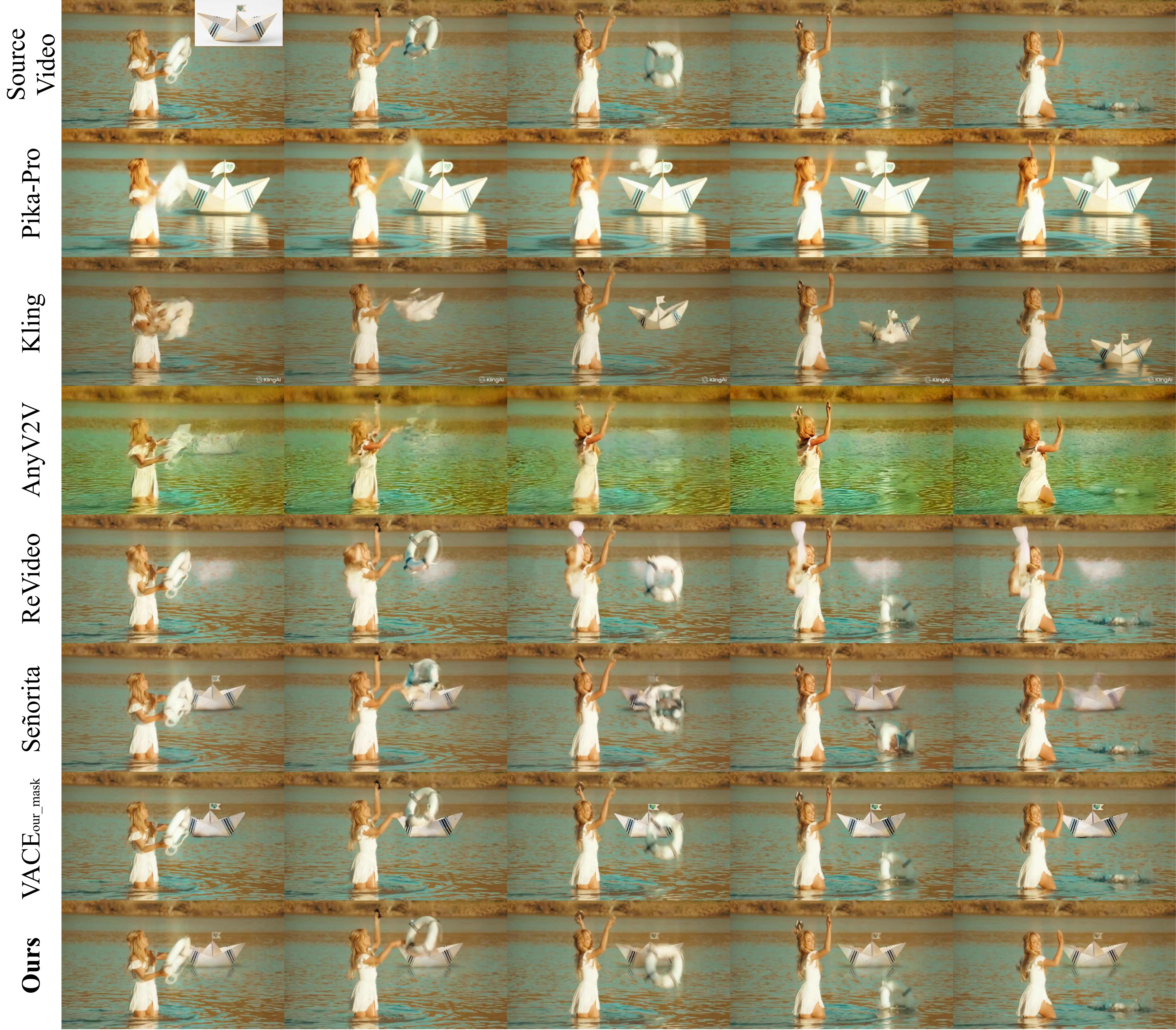}
\caption{Additional qualitative results. (Kling / Pika-Pro prompt: ``Add a paper boat to the right rear of the river in the video background.'')}
    \label{fig:sabrina}
\end{figure}

\begin{figure*}[t]
    \centering
    \includegraphics[width=1.0\textwidth]{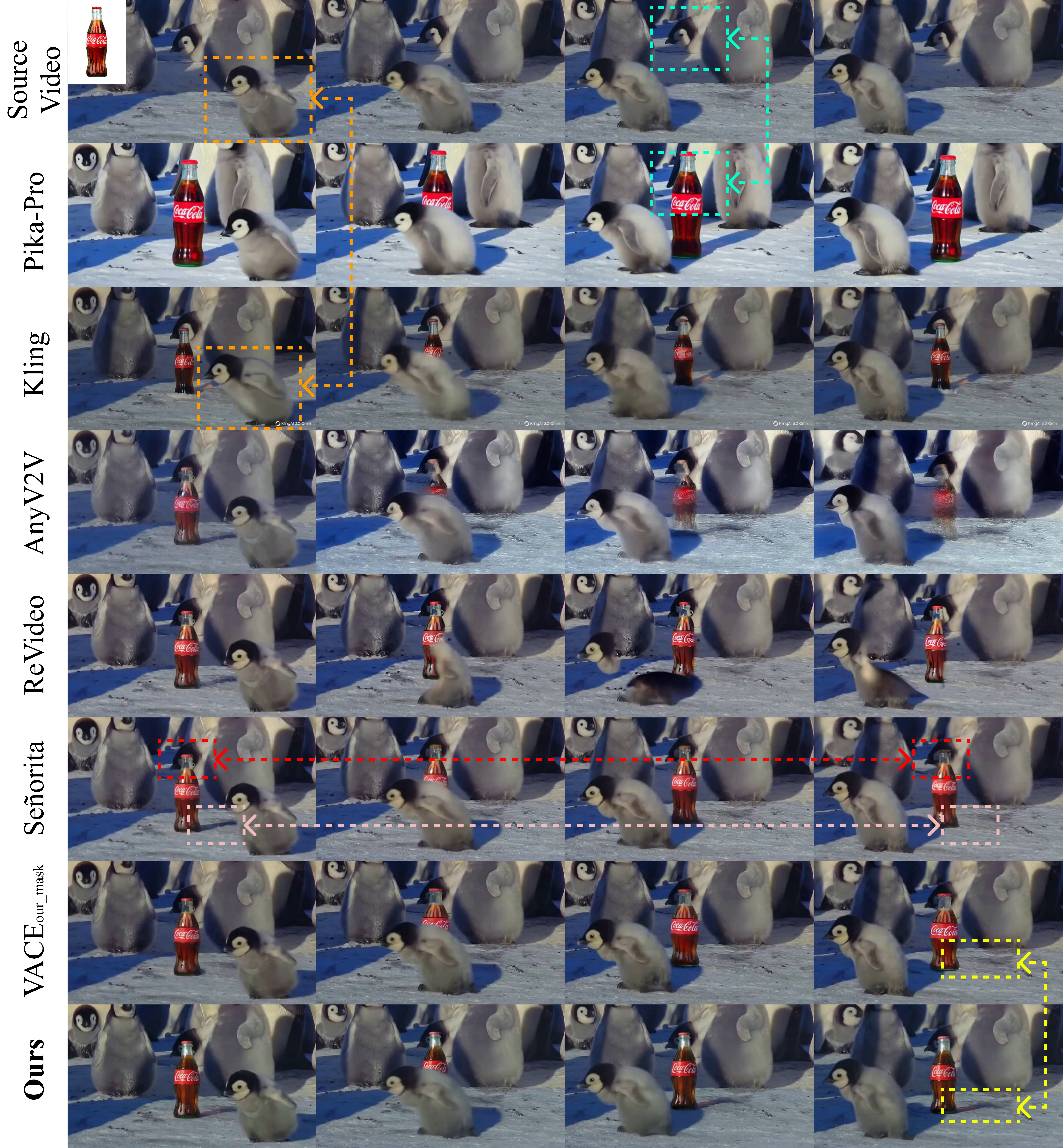}
\caption{Additional qualitative results. (Kling / Pika-Pro prompt: ``Add a coca cola behind the moving penguin in video.'')}
    \label{fig:penguin}
\end{figure*}

\begin{figure*}[t]
    \centering
    \includegraphics[width=1.0\textwidth]{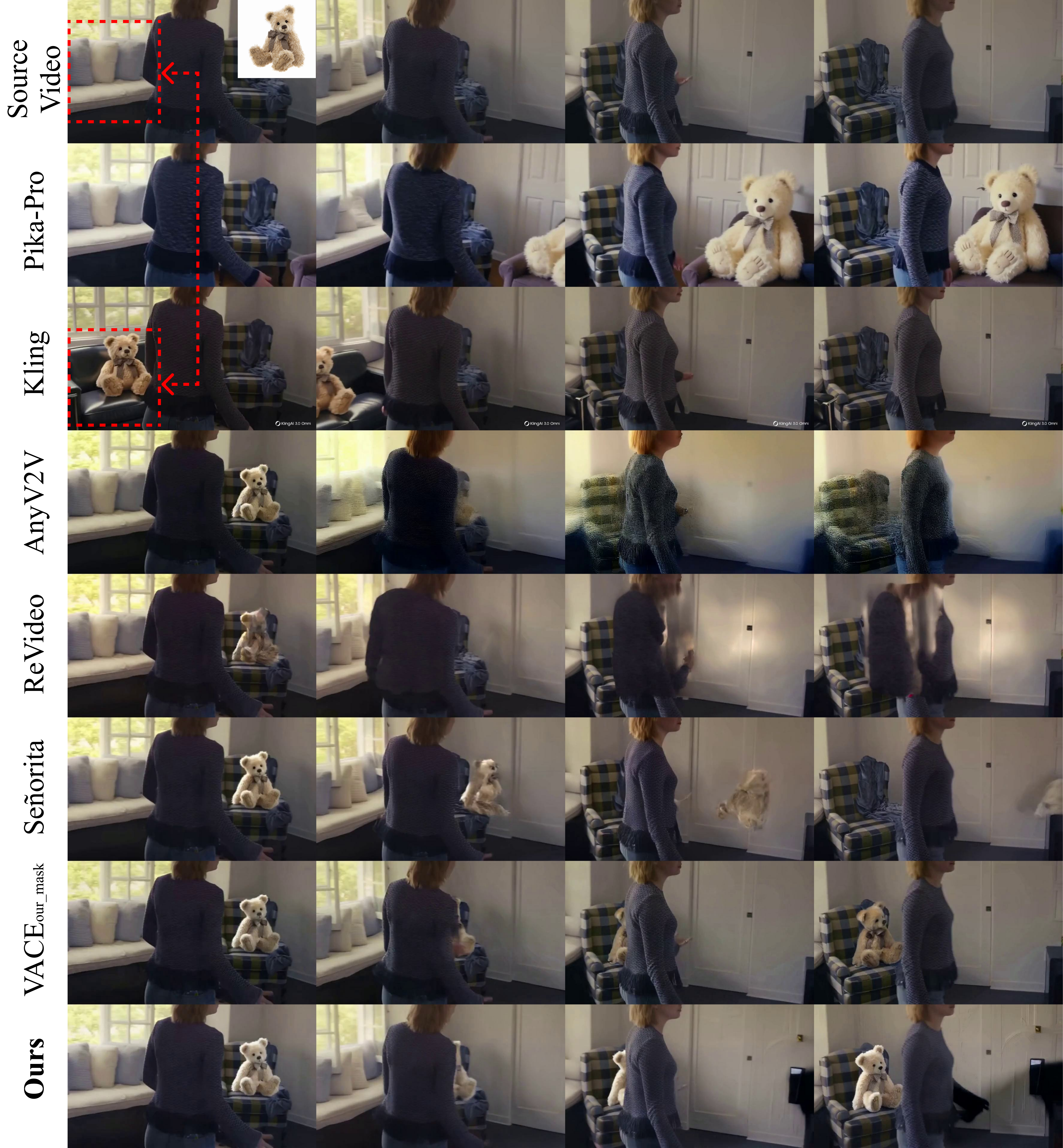}
\caption{Additional qualitative results. (Kling / Pika-Pro prompt: ``Add a teddy bear right sofa in video background.'')}
    \label{fig:emma}
\end{figure*}

\begin{figure*}[t]
    \centering
    \includegraphics[width=1.0\textwidth]{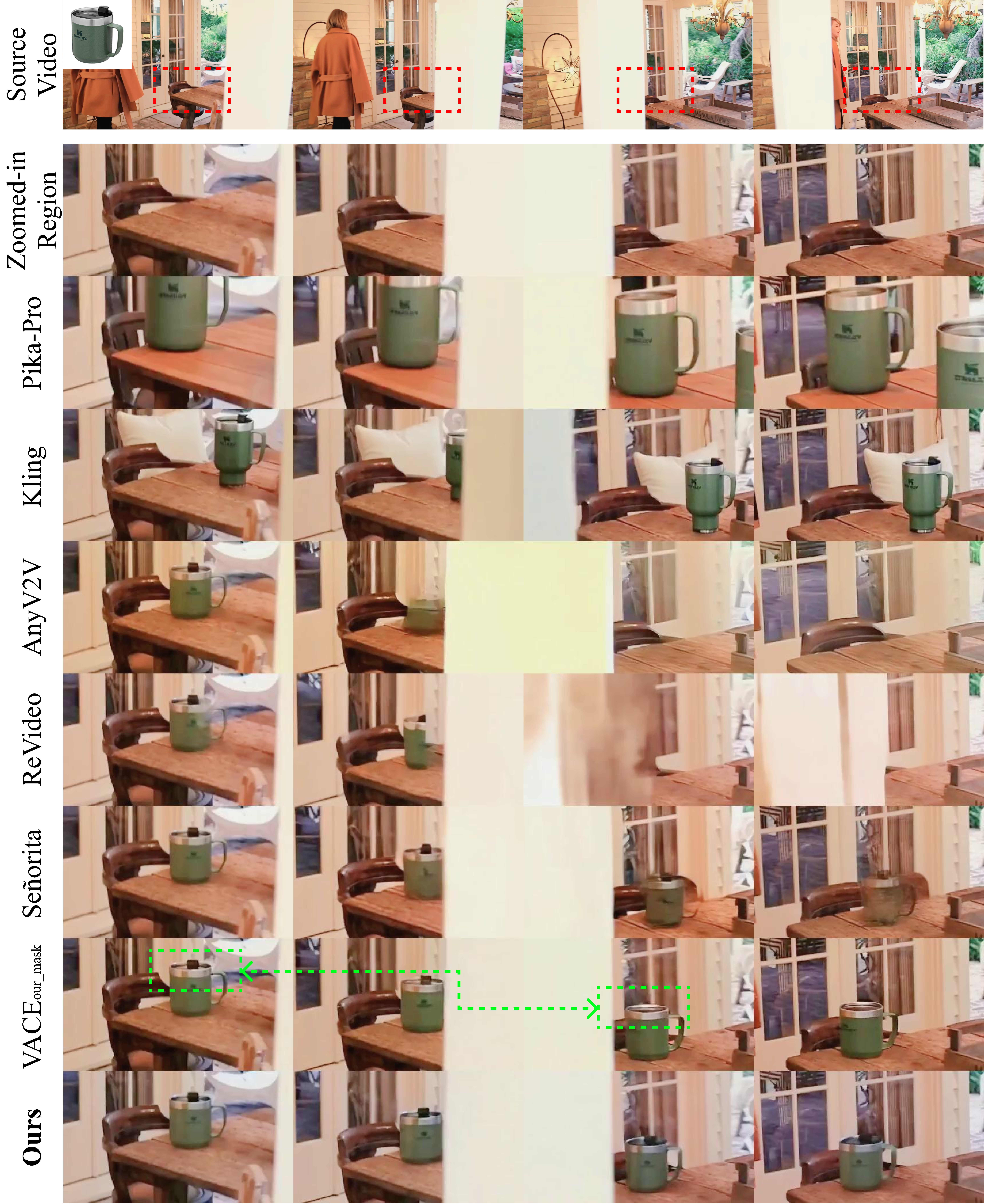}
\caption{Additional qualitative results. (Kling / Pika-Pro prompt: ``Add a stanley mug on the table back of pillar in video.'')}
    \label{fig:stanley}
\end{figure*}

Fig.~\ref{fig:sabrina},~\ref{fig:penguin},~\ref{fig:emma} and~\ref{fig:stanley} present additional qualitative results of VOI in real-world scenarios.
We compare our method against two commercial models, Pika-Pro~\cite{pikaadditions2025} and Kling~\cite{Kling}, and four open-source approaches: AnyV2V~\cite{ku2024anyv2v}, ReVideo~\cite{mou2024revideo}, Señorita~\cite{zi2025senorita}, and VACE~\cite{jiang2025vace}.
Given the varying input modalities of these models, we adapt the evaluation setup to ensure a fair comparison.
Since Pika-Pro and Kling do not support mask inputs, we use text prompts to describe the target insertion areas as precisely as possible; the specific prompts used are detailed in each figure's caption.
For AnyV2V, ReVideo, and Señorita, the models do not support mask or reference image inputs.
Instead, they accept a source video paired with an edited first frame; following their intended pipeline, we provide the source video and the same edited first frame used by our method.
Finally, for VACE, which serves as our primary baseline and supports all relevant inputs, we provide the source video and reference image, along with our 4D-aware mask sequence.
To ensure a fair comparison, we additionally supply VACE with the same edited first frame used by our method.

As shown in the figures, Pika-Pro and Kling tend to modify the original scene rather than performing localized insertion. 
For instance, in Fig.~\ref{fig:penguin}, Pika-Pro removes the penguin lying behind the insertion region.
Similarly, Kling swaps the existing tube with the target object in Fig.~\ref{fig:sabrina}, while in Fig.~\ref{fig:emma}, it removes the original sofa and generates a new black sofa instead.
In addition, Pika-Pro frequently alters the global appearance of the video, including brightness and color tone.
These models also struggle to faithfully preserve the appearance of the given object: Pika-Pro distorts the flag details of the paper boat in Fig.~\ref{fig:sabrina}, while Kling generates a mug with a different shape from the reference in Fig.~\ref{fig:stanley}.

For AnyV2V, ReVideo, and Señorita, the main limitation appears in scenes involving occlusion.
Across all four figures, these methods either fail to preserve the original content of the source video or fail to generate the target object accurately under occlusion, highlighting their limited ability to handle complex spatial relationships.

VACE, benefiting from our 4D mask as input, produces the most stable results among the baselines. 
However, it still occasionally distorts the appearance of the inserted object. 
Most notably, VACE completely fails to synthesize side effects such as shadows and reflections, which are critical for achieving realistic and coherent object insertion.

In contrast, our InsertAnywhere preserves the original scene content while accurately inserting the reference object into the target region, and naturally synthesizes optical effects such as shadows and reflections, resulting in more realistic and coherent compositions.

\begin{figure}[t]
    \centering
    \includegraphics[width=1.0\textwidth]{figures/2026eccv_supple_4dmask.pdf}
    \caption{Visualization of 4D-aware mask sequences generated by our geometrically grounded mask propagation module.}
    \label{fig:4d_mask}
\end{figure}

Furthermore, Fig.~\ref{fig:4d_mask} visualizes the 4D-aware mask sequences generated by our geometrically grounded mask propagation module.
Our masks are temporally propagated according to the reconstructed 4D scene geometry and camera motion, maintaining consistent alignment with the insertion region across frames.
This temporally coherent guidance provides a more reliable conditioning signal for the generative model during video synthesis.

%% file: main.bib
@String(NeurIPS = {Adv. Neural Inform. Process. Syst.})

@String(ICLR  = {Int. Conf. Learn. Represent.})

@String(NeurIPS = {NeurIPS})

@String(ICLR  = {ICLR})

@misc{Kling,
  title = {KlingAI},
  author = {Kling AI Team},
  url={https://app.klingai.com/global/multimodal-to-video/add-object/new},
  year = {2025}
}

@misc{pikaadditions2025,
  author       = {Pika},
  title        = {Pika Additions},
  howpublished = {\url{https://pika.art/pikadditions}},
  year         = {2025},
  note         = {Accessed: 2025-11-14}
}

@article{jiang2025vace,
  title={Vace: All-in-one video creation and editing},
  author={Jiang, Zeyinzi and Han, Zhen and Mao, Chaojie and Zhang, Jingfeng and Pan, Yulin and Liu, Yu},
  journal={arXiv preprint arXiv:2503.07598},
  year={2025}
}

@article{ku2024anyv2v,
  title={Anyv2v: A tuning-free framework for any video-to-video editing tasks},
  author={Ku, Max and Wei, Cong and Ren, Weiming and Yang, Harry and Chen, Wenhu},
  journal={Transactions on Machine Learning Research},
  year={2024}
}

@article{mou2024revideo,
  title={Revideo: Remake a video with motion and content control},
  author={Mou, Chong and Cao, Mingdeng and Wang, Xintao and Zhang, Zhaoyang and Shan, Ying and Zhang, Jian},
  journal={Advances in Neural Information Processing Systems},
  volume={37},
  pages={18481--18505},
  year={2024}
}

@inproceedings{zi2025senorita, 
  title={Señorita-2M: A High-Quality Instruction-based Dataset for General Video Editing by Video Specialists}, 
  author={Bojia Zi and Penghui Ruan and Marco Chen and Xianbiao Qi and Shaozhe Hao and Shihao Zhao and Youze Huang and Bin Liang and Rong Xiao and Kam-Fai Wong}, 
  booktitle={NeurIPS D\&B},
  year={2025}, 
}

@inproceedings{chen2024anydoor,
  title={Anydoor: Zero-shot object-level image customization},
  author={Chen, Xi and Huang, Lianghua and Liu, Yu and Shen, Yujun and Zhao, Deli and Zhao, Hengshuang},
  booktitle={Proceedings of the IEEE/CVF conference on computer vision and pattern recognition},
  pages={6593--6602},
  year={2024}
}

@inproceedings{tu2025videoanydoor,
  title={Videoanydoor: High-fidelity video object insertion with precise motion control},
  author={Tu, Yuanpeng and Luo, Hao and Chen, Xi and Ji, Sihui and Bai, Xiang and Zhao, Hengshuang},
  booktitle={Proceedings of the Special Interest Group on Computer Graphics and Interactive Techniques Conference Conference Papers},
  pages={1--11},
  year={2025}
}

@article{chen2025omniinsert,
  title={OmniInsert: Mask-Free Video Insertion of Any Reference via Diffusion Transformer Models},
  author={Chen, Jinshu and Li, Xinghui and Bai, Xu and Ma, Tianxiang and Zhang, Pengze and Chen, Zhuowei and Li, Gen and Liu, Lijie and Zhao, Songtao and Li, Bingchuan and others},
  journal={arXiv preprint arXiv:2509.17627},
  year={2025}
}

@article{anything,
  title={Anything in any scene: Photorealistic video object insertion},
  author={Bai, Chen and Shao, Zeman and Zhang, Guoxiang and Liang, Di and Yang, Jie and Zhang, Zhuorui and Guo, Yujian and Zhong, Chengzhang and Qiu, Yiqiao and Wang, Zhendong and others},
  journal={arXiv preprint arXiv:2401.17509},
  year={2024}
}

@article{invi,
  title={Invi: Object insertion in videos using off-the-shelf diffusion models},
  author={Saini, Nirat and Bodla, Navaneeth and Shrivastava, Ashish and Ravichandran, Avinash and Zhang, Xiao and Shrivastava, Abhinav and Singh, Bharat},
  journal={arXiv preprint arXiv:2407.10958},
  year={2024}
}

@inproceedings{genprop,
  title={Generative video propagation},
  author={Liu, Shaoteng and Wang, Tianyu and Wang, Jui-Hsien and Liu, Qing and Zhang, Zhifei and Lee, Joon-Young and Li, Yijun and Yu, Bei and Lin, Zhe and Kim, Soo Ye and others},
  booktitle={Proceedings of the Computer Vision and Pattern Recognition Conference},
  pages={17712--17722},
  year={2025}
}

@article{ravi2024sam,
  title={Sam 2: Segment anything in images and videos},
  author={Ravi, Nikhila and Gabeur, Valentin and Hu, Yuan-Ting and Hu, Ronghang and Ryali, Chaitanya and Ma, Tengyu and Khedr, Haitham and R{\"a}dle, Roman and Rolland, Chloe and Gustafson, Laura and others},
  journal={arXiv preprint arXiv:2408.00714},
  year={2024}
}

@inproceedings{yao2025uni4d,
  title={Uni4D: Unifying Visual Foundation Models for 4D Modeling from a Single Video},
  author={Yao, David Yifan and Zhai, Albert J and Wang, Shenlong},
  booktitle={Proceedings of the Computer Vision and Pattern Recognition Conference},
  pages={1116--1126},
  year={2025}
}

@article{piccinelli2025unidepthv2,
  title={Unidepthv2: Universal monocular metric depth estimation made simpler},
  author={Piccinelli, Luigi and Sakaridis, Christos and Yang, Yung-Hsu and Segu, Mattia and Li, Siyuan and Abbeloos, Wim and Van Gool, Luc},
  journal={arXiv preprint arXiv:2502.20110},
  year={2025}
}

@inproceedings{karaev2025cotracker3,
  title={Cotracker3: Simpler and better point tracking by pseudo-labelling real videos},
  author={Karaev, Nikita and Makarov, Yuri and Wang, Jianyuan and Neverova, Natalia and Vedaldi, Andrea and Rupprecht, Christian},
  booktitle={Proceedings of the IEEE/CVF International Conference on Computer Vision},
  pages={6013--6022},
  year={2025}
}

@inproceedings{kirillov2023segment,
  title={Segment anything},
  author={Kirillov, Alexander and Mintun, Eric and Ravi, Nikhila and Mao, Hanzi and Rolland, Chloe and Gustafson, Laura and Xiao, Tete and Whitehead, Spencer and Berg, Alexander C and Lo, Wan-Yen and others},
  booktitle={Proceedings of the IEEE/CVF international conference on computer vision},
  pages={4015--4026},
  year={2023}
}

@inproceedings{liu2024grounding,
  title={Grounding dino: Marrying dino with grounded pre-training for open-set object detection},
  author={Liu, Shilong and Zeng, Zhaoyang and Ren, Tianhe and Li, Feng and Zhang, Hao and Yang, Jie and Jiang, Qing and Li, Chunyuan and Yang, Jianwei and Su, Hang and others},
  booktitle={European conference on computer vision},
  pages={38--55},
  year={2024},
  organization={Springer}
}

@inproceedings{cheng2023tracking,
  title={Tracking anything with decoupled video segmentation},
  author={Cheng, Ho Kei and Oh, Seoung Wug and Price, Brian and Schwing, Alexander and Lee, Joon-Young},
  booktitle={Proceedings of the IEEE/CVF International Conference on Computer Vision},
  pages={1316--1326},
  year={2023}
}

@inproceedings{xiang2025structured,
  title={Structured 3d latents for scalable and versatile 3d generation},
  author={Xiang, Jianfeng and Lv, Zelong and Xu, Sicheng and Deng, Yu and Wang, Ruicheng and Zhang, Bowen and Chen, Dong and Tong, Xin and Yang, Jiaolong},
  booktitle={Proceedings of the Computer Vision and Pattern Recognition Conference},
  pages={21469--21480},
  year={2025}
}

@inproceedings{wang2024sea,
  title={Sea-raft: Simple, efficient, accurate raft for optical flow},
  author={Wang, Yihan and Lipson, Lahav and Deng, Jia},
  booktitle={European Conference on Computer Vision},
  pages={36--54},
  year={2024},
  organization={Springer}
}

@article{vbench++,
  title={Vbench++: Comprehensive and versatile benchmark suite for video generative models},
  author={Huang, Ziqi and Zhang, Fan and Xu, Xiaojie and He, Yinan and Yu, Jiashuo and Dong, Ziyue and Ma, Qianli and Chanpaisit, Nattapol and Si, Chenyang and Jiang, Yuming and others},
  journal={arXiv preprint arXiv:2411.13503},
  year={2024}
}

@article{miao2025rose,
  title={ROSE: Remove Objects with Side Effects in Videos},
  author={Miao, Chenxuan and Feng, Yutong and Zeng, Jianshu and Gao, Zixiang and Liu, Hantang and Yan, Yunfeng and Qi, Donglian and Chen, Xi and Wang, Bin and Zhao, Hengshuang},
  journal={arXiv preprint arXiv:2508.18633},
  year={2025}
}

@inproceedings{clip,
  title={Learning transferable visual models from natural language supervision},
  author={Radford, Alec and Kim, Jong Wook and Hallacy, Chris and Ramesh, Aditya and Goh, Gabriel and Agarwal, Sandhini and Sastry, Girish and Askell, Amanda and Mishkin, Pamela and Clark, Jack and others},
  booktitle={International conference on machine learning},
  pages={8748--8763},
  year={2021},
  organization={PMLR}
}

@article{dinov2,
  title={Dinov2: Learning robust visual features without supervision},
  author={Oquab, Maxime and Darcet, Timoth{\'e}e and Moutakanni, Th{\'e}o and Vo, Huy and Szafraniec, Marc and Khalidov, Vasil and Fernandez, Pierre and Haziza, Daniel and Massa, Francisco and El-Nouby, Alaaeldin and others},
  journal={arXiv preprint arXiv:2304.07193},
  year={2023}
}

@article{zhao2025dreaminsert,
  title={DreamInsert: Zero-Shot Image-to-Video Object Insertion from A Single Image},
  author={Zhao, Qi and Ma, Zhan and Zhou, Pan},
  journal={arXiv preprint arXiv:2503.10342},
  year={2025}
}

@article{hu2022lora,
  title={Lora: Low-rank adaptation of large language models},
  author={Hu, Edward J and Shen, Yelong and Wallis, Phillip and Allen-Zhu, Zeyuan and Li, Yuanzhi and Wang, Shean and Wang, Lu and Chen, Weizhu and others},
  journal={ICLR},
  year={2022}
}

@article{yu2025omnipaint,
  title={Omnipaint: Mastering object-oriented editing via disentangled insertion-removal inpainting},
  author={Yu, Yongsheng and Zeng, Ziyun and Zheng, Haitian and Luo, Jiebo},
  journal={arXiv preprint arXiv:2503.08677},
  year={2025}
}

@article{achiam2023gpt,
  title={Gpt-4 technical report},
  author={Achiam, Josh and Adler, Steven and Agarwal, Sandhini and Ahmad, Lama and Akkaya, Ilge and Aleman, Florencia Leoni and Almeida, Diogo and Altenschmidt, Janko and Altman, Sam and Anadkat, Shyamal and others},
  journal={arXiv preprint arXiv:2303.08774},
  year={2023}
}
